\documentclass[lettersize,journal]{IEEEtran}
\usepackage{amsmath,amsfonts}
\usepackage{algorithmic}
\usepackage{array}
\usepackage{stfloats}
\usepackage{url}
\usepackage{verbatim}
\usepackage{graphicx}
\usepackage{float}
\usepackage[export]{adjustbox} 
\usepackage{multirow}
\usepackage[T1]{fontenc}
\usepackage{booktabs}
\usepackage{multirow}
\usepackage{makecell}
\usepackage{mathabx}
\usepackage{amsmath}
\usepackage{subfigure}
\usepackage{verbatim}
\usepackage{cite}
\usepackage{afterpage}
\usepackage{pifont}
\usepackage{balance}
\usepackage[ruled,linesnumbered]{algorithm2e}
\usepackage[colorlinks,
linkcolor=blue,
anchorcolor=blue,
citecolor=blue]{hyperref}
\usepackage[table]{xcolor}

\definecolor{tabfirst}{rgb}{1, 0.75, 0.7}
\definecolor{tabsecond}{rgb}{1, 0.83, 0.7}
\definecolor{tabthird}{rgb}{1, 0.96, 0.7}

\usepackage{threeparttable}

\def\BibTeX{{\rm B\kern-.05em{\sc i\kern-.025em b}\kern-.08em
    T\kern-.1667em\lower.7ex\hbox{E}\kern-.125emX}}
\usepackage{balance}

\title{Dream-SLAM: Dreaming the Unseen for \\ Active SLAM in Dynamic Environments}
\author{Xiangqi Meng, Pengxu Hou, Zhenjun Zhao, Javier Civera, Daniel Cremers, Hesheng Wang, Haoang Li
\thanks{X. Meng, P. Hou, and H. Li are with the Thrust of Robotics and Autonomous Systems, Hong Kong University of Science and Technology (Guangzhou), Guangzhou, China.}
\thanks{
Z. Zhao and J. Civera are with the University of Zaragoza, Zaragoza, Spain.
}
\thanks{
D. Cremers is with the 
School of Computation, Information and Technology, 
Technical University of Munich, Munich, Germany.
}
\thanks{
	H. Wang is with the School of Automation and IntelligentSensing, Shanghai Jiao Tong University, Shanghai 200240, China (e-mail: wanghesheng @sjtu.edu.cn).
}
}

\begin{document}

\maketitle

\begin{abstract}
In addition to the core tasks of simultaneous localization and mapping (SLAM), active SLAM additionally involves 
generating robot actions that
enable effective and efficient exploration of unknown environments.
However, existing active SLAM pipelines are limited by three main factors. First, they inherit the 
restrictions
of the underlying SLAM modules that they may be using. Second, their motion planning strategies are typically shortsighted and lack long-term vision. Third, most approaches struggle to handle dynamic scenes.
To address these limitations, 
we propose a novel monocular active SLAM method, Dream-SLAM, which is based on \emph{dreaming} 
cross-spatio-temporal images
and semantically plausible structures of partially observed dynamic environments.
The generated cross-spatio-temporal images are fused with real observations to mitigate noise and data incompleteness, leading to more accurate camera pose estimation and a more coherent 3D scene representation. Furthermore, we integrate dreamed and observed scene structures to enable long-horizon planning, producing farsighted trajectories that promote efficient and thorough exploration.
Extensive experiments on both public and self-collected datasets demonstrate that Dream-SLAM outperforms state-of-the-art methods in localization accuracy, mapping quality, and exploration efficiency. Source code will be publicly available upon paper acceptance.

\end{abstract}


\section{Introduction}
Simultaneous Localization and Mapping (SLAM) addresses the joint estimation of the ego-pose of a moving platform and a representation of its surrounding environment, and constitutes a fundamental building block for a wide range of emerging applications, including virtual, augmented, and mixed reality, as well as autonomous robots. In its most general formulation, the SLAM estimates are passively updated using measurements acquired from the platform’s onboard sensors~\cite{cadena2016past}, which may limit the accuracy, completeness, and task-specific relevance of the resulting map. In contrast, robotic platforms can actively control their motion to deliberately gather informative measurements that improve the quality of the estimated representation. This paradigm, commonly referred to as active SLAM~\cite{placed2023survey}, has enabled significant advances in applications such as search-and-rescue operations, warehouse inventory management, and large-scale scene reconstruction~\cite{brugali2025mobile}.

Despite remarkable progress in recent years, active SLAM remains constrained by three major limitations. First, 
its overall performance is strongly conditioned by the quality of the underlying estimation modules. Most research, however, focuses primarily on the action or planning components, while relying on off-the-shelf localization and mapping strategies~\cite{cao2021tare,yu2023echo}, or evaluating exploration performance using ground-truth camera poses~\cite{ActiveSplat,yan2023active}. Since localization and mapping constitute the backbone of active SLAM, improvements in these components directly translate into gains across the entire pipeline. Second, from a planning perspective, the majority of methods adopt either frontier-based~\cite{FBE} or sampling-based~\cite{7487281} strategies.
While effective in certain scenarios, these planners are inherently shortsighted: they operate solely on the currently observed map and lack mechanisms to reason about or anticipate unexplored regions. As a consequence, they often converge to locally optimal trajectories with unnecessary detours or frequent backtracking. Third, most methods~\cite{vins2018robust,dso} assume a static environment, which rarely holds in practice. In dynamic scenes, such as homes with moving occupants or crowded shopping malls, the presence of dynamic objects introduces occlusions and induces localization drift, both of which significantly degrade planning reliability and overall performance.

In recent years, there have been efforts to mitigate the aforementioned limitations. First, for more accurate localization and mapping, some methods~\cite{9636611,naveed2025help} encourage robots to preferentially move to areas with rich visual textures. This strategy is motivated by the observation that textured areas often provide distinctive and reliable features useful for camera pose estimation and 3D reconstruction. However, prioritizing highly textured areas may introduce exploration bias, leading to incomplete environment coverage as low-texture areas may be systematically neglected. 
 Second, the shortsighted planning can be alleviated by reasoning over unknown space to generate global coverage paths~\cite{Cherie2025MapEx,10038280}. However, these approaches often assume that unexplored regions exhibit regular layouts, resulting in an overly simplified approximation. Third, to tackle dynamic scenes, several methods~\cite{zhang2020vdo,jiang2024rodyn} explicitly filter out moving objects to maintain a static background map, discarding informative foreground content. More recent work~\cite{PG-SLAM} can reconstruct the dynamic foreground, but tends to become unstable under fast or complex motions, and also significantly increases system complexity. 

\begin{figure*}[!t] 
\centering
    \includegraphics[width=1.0\textwidth]{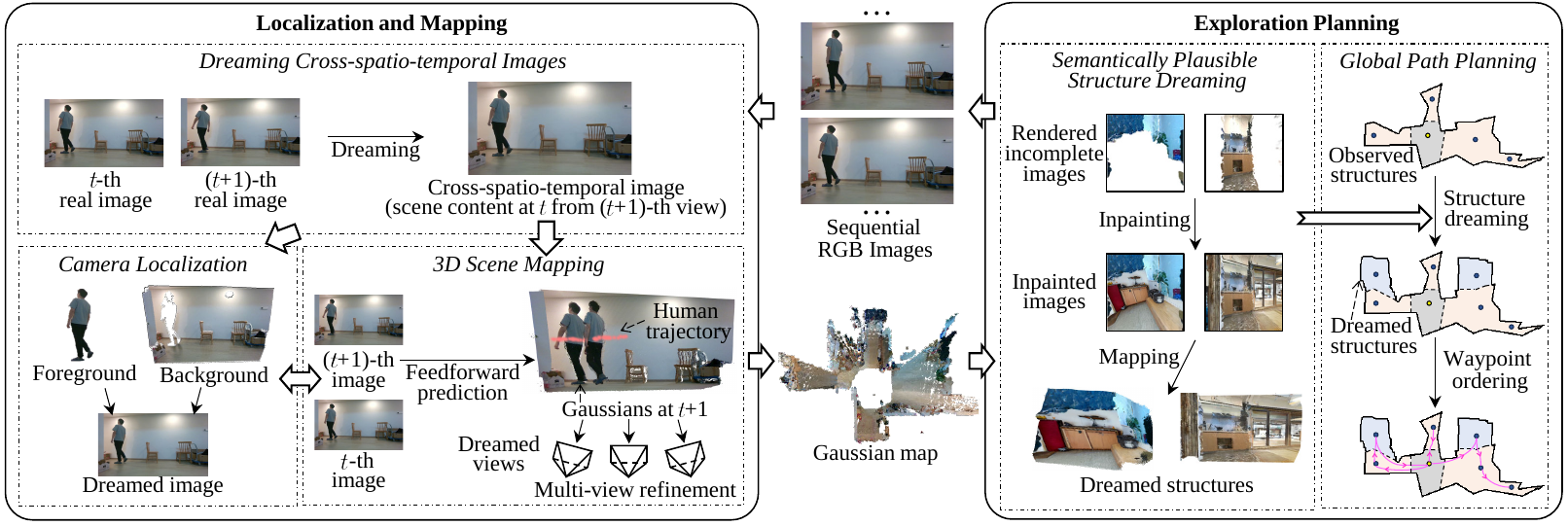} 
    \caption{\textbf{Dream-SLAM overview.} Our pipeline consists of two main modules: localization and mapping, and exploration planning. (a) For localization, we propose to dream cross-spatio-temporal images, and use these images to construct additional 3D-2D foreground constraints that can effectively compensate for noise. For mapping, we propose a feedforward network to reconstruct per-pixel Gaussians of both static background and dynamic foreground. 
    We further refine Gaussians based on multi-view constraints provided by cross-spatio-temporal and real images.
    (b) Our planning module dreams semantically plausible structures of unobserved areas. By integrating the dreamed and observed information, we plan a 
    farsighted
    path, enabling an efficient and thorough exploration.}
    \label{fig:overview}
\end{figure*}

To address the above challenges, we propose a monocular active SLAM method, named \textbf{Dream-SLAM}, via dreaming the unseen in dynamic environments.
The core idea of our method is dreaming the cross-spatio-temporal images
and the semantically plausible structures of the scene.
For one thing, the cross-spatio-temporal images can be combined with real images to compensate for noise and data incompleteness. This combination leads to a more 
accurate camera pose estimation, as well as a more 
coherent 3D scene representation.
For another, 
we integrate the dreamed and observed scene structures for long-horizon planning. This way generates a 
farsighted path that can efficiently achieve a thorough exploration. As shown in Fig.~\ref{fig:overview}, our method is composed of two main modules for localization and mapping, as well as exploration planning. The localization and mapping module
supplies the observed environment information to the planning module. In turn, the planner provides motion commands that drive the robot to acquire new observations.

As to camera localization in dynamic environments, a common practice is to use 3D-2D constraints of the static background only. While the dynamic foreground has the potential to improve the accuracy, it can hardly be used in practice. The reason is that 2D image of the foreground at the current time is inconsistent with 3D foreground reconstructed at the previous time, due to object movement. To solve this problem, we propose to dream a cross-spatio-temporal image, which depicts the dynamic 3D scene at the previous time, from the viewpoint of the current camera. 
Based on the consistent 3D-2D foreground content at the previous time, we can establish constraints on the pose of the current camera. We dream the image 
by designing a network based on diffusion~\cite{rombach2022high}. The dreamed image
exhibits high appearance fidelity and spatial rationality. In addition, to enable photo-realistic and efficient mapping of dynamic scenes, we propose a feedforward network compatible with Gaussian splatting~\cite{kerbl3Dgaussians}. This network can directly predict per-pixel Gaussians of both dynamic foreground and static background. It is more concise and general than existing Gaussian estimation solutions~\cite{WildGS,DG-SLAM}. To further refine Gaussians, we leverage both dreamed cross-spatio-temporal and real images. The dreamed views can effectively supplement the real observations by providing additional supervisory signals.

In terms of planning, we aim to overcome the shortsightedness of conventional planners~\cite{ActiveSplat,jiang2024fisherrf}. First, on the waypoints that the robot will potentially visit, we place multiple virtual cameras. We use these cameras to render the reconstructed 3D scenes, obtaining a set of incomplete images. Then we introduce a diffusion-based model to inpaint these images, generating virtual observations of the unexplored region. We further back-project the inpainted images into 3D Gaussians based on the above feedforward network. By integrating these Gaussians with the existing map, we obtain a more complete scene structure. The above dreaming process considers the observed information to infer the unseen content. Accordingly, the dreamed structures implicitly encode the information of the surroundings, and thus are semantically plausible. Planning within such  structures leads to a farsighted path. Please note that the dreamed structures are not used for final mapping, but are replaced by the real observations once visited.

Our main contributions are summarized as follows:

\begin{itemize}

\item We introduce dreaming as a unified mechanism for localization, mapping, and exploration planning. The dreamed content effectively supplements the real observations, enhancing the performance of active SLAM.
 
\item For localization, we propose to dream cross-spatio-temporal images. These images can incorporate the information of the dynamic foreground to compensate for noise and data incompleteness, improving the accuracy of camera pose estimation.

\item For mapping, we propose a feedforward network to efficiently predict Gaussians,  achieving a photo-realistic scene reconstruction. Then we refine Gaussians by the cross-spatio-temporal images, which can supplement real views and achieve a more coherent 3D representation.

\item For planning, we propose to dream the semantically plausible structures of unexplored regions. By integrating the dreamed and observed structures, our method generates a farsighted path that can efficiently achieve a thorough and efficient exploration.

\end{itemize}

Extensive experiments on both public and self-collected datasets demonstrate the superior performance of our Dream-SLAM in localization and mapping accuracy, as well as exploration efficiency, compared to state-of-the-art approaches.

\section{Related Work}
\label{sec:rel_work}

\subsection{Passive SLAM}
\textbf{Geometric Representations.}
Dominant methods estimate camera poses by establishing point correspondences across views and reconstruct sparse 3D point clouds through triangulation~\cite{mur2015orb,MonoSlam}. Some methods 
extend the sparse feature points to all the image pixels, optimizing intensity and depth residuals to achieve dense scene representation~\cite{whelan2016elasticfusion,6696650}. In recent years, several methods have incorporated deep neural networks to further enhance performance. For example, DROID-SLAM~\cite{teed2021droid} employs a recurrent network to enhance camera pose estimation through jointly learned depth and correspondence cues.
The above methods operate well in static environments, but their performance degrades when dynamic objects appear. To cope with dynamic scenes, some approaches directly detect and filter out foregrounds. Among them, DynaSLAM~\cite{bescos2018dynaslam} and ReFusion~\cite{ReFusion} utilize geometric residual, FlowFusion~\cite{zhang2020flowfusion} relies on optical flow estimation, and DS-SLAM~\cite{yu2018ds} leverages semantic segmentation. The common limitation of these approaches is that they fail to reconstruct the dynamic foregrounds.
To address this limitation, several methods have been proposed. VDO-SLAM~\cite{zhang2020vdo} models the motion of rigid items by incorporating geometric constraints. AirDOS~\cite{qiu2022airdos} and Body-SLAM~\cite{henning2022bodyslam} further model non-rigid human motion and reconstruct articulated skeletons. MonST3R~\cite{zhang2024monst3r} leverages the learned multi-view geometry prior to jointly estimate camera poses and dense 3D structure across views.
Despite these advances, they cannot reconstruct dynamic objects with photo-realistic details.

\textbf{Neural Representations.}
Neural representation, which typically includes implicit neural radiance
fields~\cite{mildenhall2020nerf} and explicit Gaussian splatting~\cite{kerbl3Dgaussians}, has recently shown strong performance in photo-realistic mapping~\cite{deng2025best}. Early works focused primarily on the use of photometric constraints in static environments~\cite{johari2023eslam,sucar2021imap,GS-SLAM,gaussianslam}. Recently, some works have extended these techniques to handle dynamic scenes. Rodyn-SLAM~\cite{jiang2024rodyn} and DG-SLAM~\cite{DG-SLAM} leverage optical flow and semantic information to estimate the mask of  dynamic objects and filter them out. WildGS-SLAM~\cite{WildGS} employs a network to learn uncertainty online, but this increases the computational complexity. These methods also cannot reconstruct the dynamic foregrounds, thereby neglecting potentially useful information.
In contrast, PG-SLAM~\cite{PG-SLAM} leverages shape priors to reconstruct both rigid and non-rigid objects, while jointly using foreground and background information to localize the camera.
However, in highly dynamic scenes, foreground reconstruction can be unreliable, and separate reconstructions for rigid and non-rigid objects further increase system complexity.

Overall, the above classical SLAM methods neither effectively leverage the foreground information to localize the camera, nor reliably reconstruct the dynamic foreground. We address these limitations from two aspects. For one thing, we dream cross-spatio-temporal images, establishing foreground-related constraints. For another, we introduce a feedforward network that can directly
predict per-pixel Gaussians of both dynamic
foreground and static background.

\subsection{Active SLAM}

\textbf{Sampling-based Exploration.}
\label{subsubsec:sampling_explore}
The sampling-based methods generate multiple waypoints within the scene, which encode position and orientation information. Among these candidate waypoints, the robot selects the optimal one.
Bircher et al.~\cite{7487281} generate a rapidly-exploring random tree within the space, and then under a receding-horizon scheme, select the branch with the most unmapped space to explore. ANM~\cite{yan2023active} employs a neural scene representation and selects viewpoints according to the uncertainty distribution of this neural map. 
ActiveGS~\cite{Jin2025ActiveGS} reconstructs a Gaussian map, considering both rendering uncertainty and path cost. ActiveGAMER~\cite{chen2025activegamer} incorporates a dynamically updated candidate pool to manage waypoints, reducing computational cost. 
The above methods partly neglect the spatial continuity between neighboring waypoints. Accordingly, the planning is prone to getting stuck into a local optimum, resulting in redundant detours and backtracking.
To solve this problem, the recent ActiveSplat~\cite{ActiveSplat} introduces a topological graph to cluster the waypoints into multiple sub-regions. When making the next-step decision, it prioritizes the exploration completeness of each sub-region 
based on information gain and distance.

\textbf{Frontier-based Exploration.}
The frontier-based exploration was first introduced in~\cite{FBE} and later systematically formalized in~\cite{julia2012comparison}. Frontiers are defined as the boundaries between occupied and unknown regions. In this paradigm, the robot selects the closest frontier as the next exploration target. Cieslewski et al.~\cite{cieslewski2017rapid} select the frontier that is within the current field of view and impose minimal impact on the speed of flight, ensuring stable and efficient exploration at high velocities. Shen et al.~\cite{shen2012stochastic} employ a stochastic differential equation–based strategy to select the region with the strongest particle expansion as the next frontier.
FisherRF~\cite{jiang2024fisherrf} represents the scene with 3D Gaussians and selects frontier-related waypoints by maximizing Fisher information. 
The above methods commonly adopt a greedy strategy, which tends to a locally maximum information gain. To address this, some methods incorporate global information into the planning process. For example, FUEL~\cite{zhou2021fuel} adopts a hierarchical planning strategy to obtain a global exploration path. FALCON~\cite{10816079} decomposes the unknown space into multiple disjoint zones, and subsequently plans a path to effectively cover all of them. 
In addition to the observed information, recent methods~\cite{Cherie2025MapEx, 10038280} alleviate
the shortsighted planning by reasoning over unknown space. However, they often assume that unexplored regions exhibit regular layouts, resulting in an overly simplified approximation.

Overall, the above active SLAM approaches either rely solely on local observations, or infer an unrealistic approximation of unexplored areas. These limitations lead to a locally optimal path.
By contrast, our method dreams the semantically plausible structures of unexplored regions. Accordingly, our planner can reason over a more complete scene layout, producing a more farsighted
path.

\section{Problem Formulation}

\subsection{Preliminaries}
\label{sec:Preliminaries}
\textbf{Diffusion Models.} 
Diffusion model~\cite{rombach2022high} is a powerful image generation tool. The forward process gradually adds Gaussian noise to the latent code that encodes the input image, while the reverse process learns to remove this noise step by step. At a certain step~$s$, given a noisy latent code~$\mathbf{z}_s$, a network~$\mathcal{D}$ is used to predict the noise as $\epsilon = \mathcal{D}(\mathbf{z}_s, s)$. The predicted noise~$\epsilon$ is a function with respect to the network~$\mathcal{D}$. We can train~$\mathcal{D}$ via the loss
$
        \mathcal{L} = \mathbb{E} \big[\left\| \hat{\epsilon} - \epsilon(\mathcal{D})\right\|_2\big],
$
where $\mathbb{E}$ represents the expectation, and $\hat{\epsilon}$ denotes the ground-truth added noise.

\textbf{3D Foundation Models.} 
We introduce a representative model DUSt3R~\cite{dust3r} that can produce a 3D point cloud from images in static environments. The network first 
encodes two consecutive images separately to obtain image features. These features are then fed to two branch decoders, each of which consists of multiple blocks.  Based on the cross-attention, decoders generate image tokens. Finally, a point cloud head 
 of the first branch takes corresponding tokens as input and predicts the point cloud in the first camera frame. The second branch follows the same process to obtain a point cloud in the first camera frame.

\textbf{Gaussian Splatting.}
\label{GS}
A 3D scene can be represented in a photo-realistic way through a set of 3D Gaussians
$\mathcal{G}$~\cite{kerbl3Dgaussians}. 
Each Gaussian is parameterized 
by its center, covariance matrix, opacity, and color.
Through differentiable rendering~$\pi[\cdot]$, an image can be obtained as $\tilde{I}=\pi[\mathcal{G}]$.
To optimize Gaussians~$\mathcal{G}$, we minimize the appearance difference between the rendered image~$\tilde{I}$ and the ground-truth image~$I$
based on the photometric loss:
$\mathcal{L}_{\textnormal{photo}}(I, \tilde{I}) = \alpha \cdot \|I - \tilde{I}\|_1 +
    (1-\alpha ) \cdot \big(1-\textnormal{SSIM}(I,\tilde{I}) \big)$,
where 
$\alpha$ controls the trade-off between the $L_1$ and SSIM terms~\cite{ssim}.

\subsection{Dream-SLAM Overview}

As shown in Fig.~\ref{fig:overview}, our Dream-SLAM takes RGB images as input and does not rely on depth images. It consists of two modules for
localization and mapping, as well as exploration planning. We segment dynamic objects in each image based on Mask R-CNN~\cite{maskrcnn}. Our approach can handle both rigid objects (e.g., boxes) and non-rigid objects (e.g., humans and animals). Without loss of generality, we focus on humans as the main illustrative example, while experiments on other object categories are reported in the supplementary material. Moreover, we follow~\cite{PG-SLAM} to perform bundle adjustment and loop closure for optimization.

\textbf{Localization and Mapping.} 
For localization, given two images obtained at the previous and current times, we employ a diffusion model to dream a cross-spatio-temporal image.
 Such an image depicts the previous 3D scene from the current viewpoint. Using both the foreground and background of this image as supervision, we define photometric constraints to optimize the camera pose. 
To efficiently map both foreground and background in a photo-realistic manner, we design a feedforward network that can directly estimate pixel-wise Gaussians. In addition, we optimize Gaussians using multi-view photometric constraints provided by not only the real images, but also the dreamed cross-spatio-temporal images. Details are available in Section~\ref{sec:Localization and Mapping}.
\begin{figure}[!t]
\footnotesize
\centering
\renewcommand{\tabcolsep}{0.3pt}
\begin{tabular}{cc}

\includegraphics[height=0.63\linewidth]{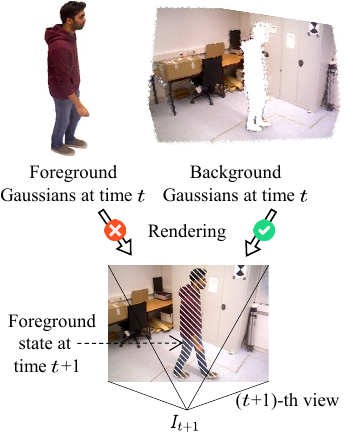}  & 
\includegraphics[height=0.63\linewidth]{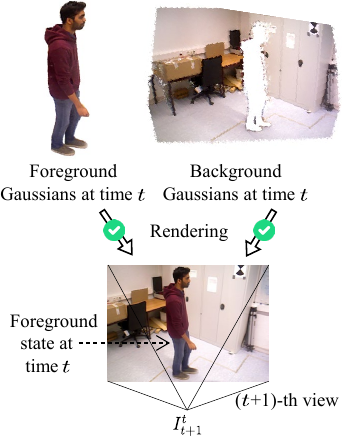} \\
(a) &  (b)
\end{tabular}
\caption{\textbf{Cross-spatio-temporal images for camera localization.}
(a) Traditional localization methods rely solely on the static background to estimate the camera pose. (b) In contrast, our method leverages both the dynamic foreground and static background by aligning the Gaussians' rendering at time~$t$ with the dreamed cross-spatio-temporal image~$I^t_{t+1}$, which represents the scene at time~$t$ from the viewpoint of the $(t+1)$-th camera.}
\label{fig:diff_camera}
\end{figure}

\textbf{Exploration Planning.}
We construct a 2D topological map from the Gaussian map described above, representing both 
waypoints and their connectivity.
The robot iteratively selects the optimal waypoint to move. Differently from existing shortsighted planners, we propose to dream semantically plausible structures of unexplored areas, followed by integrating the observed and dreamed information for long-horizon planning. Specifically, we first
employ the diffusion model to inpaint the missing content of images rendered at unvisited waypoints. Then we use a set of inpainted images to reconstruct Gaussians of the unobserved areas, and integrate them into the existing Gaussian map. Based on the enriched map, we further update the topological map, in which we plan a farsighted
path. Details are available in Section~\ref{sec:planning}.

\section{Localization and Mapping}
\label{sec:Localization and Mapping}
In this section, we introduce how we overcome the challenges raised by dynamic objects in localization and mapping. We propose to dream cross-spatio-temporal images that contribute to the accuracy improvement of localization, as well as a coherent 3D representation. Without loss of generality, let us consider the images~$I_t$ and $I_{t+1}$ obtained at times $t$ and $(t+1)$ for illustration.

\subsection{Dreaming Cross-spatio-temporal Images} 
\label{Spatio-temporal}
We begin by introducing the definition and role of cross-spatio-temporal images in the camera localization task.
As illustrated in Fig.~\ref{fig:diff_camera}(a),
Gaussian splatting-based localization methods typically estimate the pose of the image~$I_{t+1}$ by aligning it with a rendering of the Gaussians~$\mathcal{G}_t$ reconstructed up to time~$t$~\cite{GS-SLAM}. However, this strategy does not hold in the presence of dynamic foregrounds, as object states in the image $I_{t+1}$ are inconsistent with those encoded in Gaussians~$\mathcal{G}_t$.

To overcome this limitation, we propose to
dream a cross-spatio-temporal image $I^t_{t+1}$  (see Fig.~\ref{fig:diff_camera}(b)), which depicts the full scene (both dynamic foreground and static background) at time~$t$, from the viewpoint of the $(t+1)$-th camera. Accordingly, the object states in the image $I^t_{t+1}$ and Gaussians~$\mathcal{G}_t$ are consistent. We formulate the generation of $I^t_{t+1}$ as an inpainting problem in two stages, as detailed below.

\begin{figure}[!t]
\footnotesize
\centering
\includegraphics[width=\linewidth]{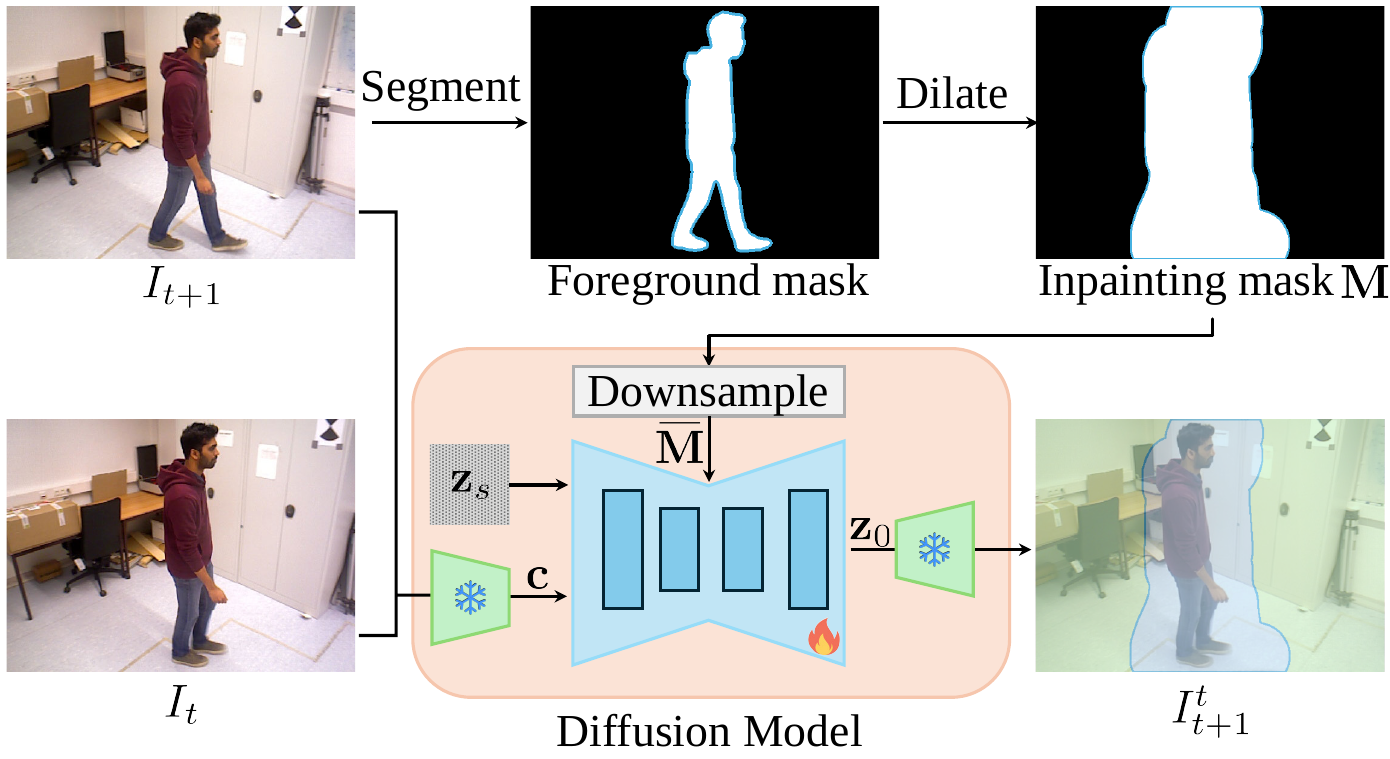}
\caption{\textbf{Dreaming a cross-spatio-temporal image. }Given the image~$I_{t+1}$, we segment the foreground and dilate the foreground mask to obtain the inpainting mask~$\mathbf{M}$. Then we feed the images $I_{t}$ and $I_{t+1}$, together with the mask~$\mathbf{M}$, into the diffusion model, which dreams the cross-spatio-temporal image~$I^t_{t+1}$.}
\label{fig:diff_image}
\end{figure}

\textbf{Generation of Inpainting Mask.}
As shown in Fig.~\ref{fig:diff_image}, the real image
$I_{t+1}$ and the cross-spatio-temporal image~$I^t_{t+1}$ share the same background, as they are rendered from the same viewpoint, but differ in their foreground due to object motion.
Accordingly, dreaming~$I^t_{t+1}$ amounts to replacing the foreground in the real image $I_{t+1}$ with the foreground state at time~$t$.
Directly using the foreground segmented from the image~$I_t$ for this replacement is, however, not appropriate due to the change in viewpoint.
Moreover, because of both object motion and viewpoint variation, parts of the foreground at time~$t$ may lie outside the foreground mask of~$I_{t+1}$. To address these challenges, we define the inpainting mask $\mathbf{M}$ of the image~$I_{t+1}$. This mask
fully encloses both the foreground to be removed in $I_{t+1}$, as well as the foreground to be fused from time ~$t$. We obtain this mask $\mathbf{M}$ by performing a lightweight dilation to the foreground mask of $I_{t+1}$. Within this mask, foreground status in $I_{t+1}$ is replaced by the foreground status at time~$t$, while the background status in $I_{t+1}$ remains unchanged. Experiments show that this mask design is robust across dynamic scenes with diverse object motions and scales.

\textbf{Two-view-guided Image Inpainting.} 
Given the images $I_t$ and $I_{t+1}$, along with the inpainting mask $\mathbf{M}$, we adapt the diffusion model introduced in Section~\ref{sec:Preliminaries} to inpaint the masked regions of $I_{t+1}$.\footnote{To match the spatial resolution of the noise used in the diffusion process, the mask $\mathbf{M}$ is downsampled, yielding $\widebar{\mathbf{M}}$.}
We first describe the forward diffusion process. We assume that the ground-truth cross-spatio-temporal image $\hat{I}_{t+1}^t$ is available (details will be introduced below).
We encode $\hat{I}_{t+1}^t$ together with the input images $I_t$ and $I_{t+1}$, using a pretrained variational autoencoder~\cite{kingma2013auto}, to obtain the latent codes $\hat{\mathbf{z}}_0$, $\mathbf{c}_1$, and $\mathbf{c}_2$, respectively. The codes~$\mathbf{c}_1$ and $\mathbf{c}_2$ are then concatenated to form a reference code~$\mathbf{c}$. Gaussian noise~$\epsilon$ is progressively added to $\hat{\mathbf{z}}_0$, yielding the noisy latent $\mathbf{z}_s$ at step~$s$.
In the reverse diffusion process, we introduce a noise prediction network~$\mathcal{D}$ to predict the added noise~$\epsilon$:
\begin{equation}
   \epsilon = \mathcal{D}(\mathbf{z}_s, s, \mathbf{c}, \widebar{\mathbf{M}}).
\end{equation}
The network~$\mathcal{D}$ is conditioned on two key inputs: the reference code $\mathbf{c}$, which encodes the unchanged context, and the inpainting mask~$\widebar{\mathbf{M}}$, which specifies the region to be inpainted. 
These conditioning signals enable context-aware inpainting, allowing the synthesized region to be seamlessly blended with its surroundings.
To train the network~$\mathcal{D}$, we use the following loss:
\begin{equation}
    \mathcal{L} = 
    \mathbb{E} \Big[\| \bar{\mathbf{M}} \odot \big(\hat{\epsilon} - \epsilon(\mathcal{D}) \big)\|_2 \Big],
\label{eq:cross_spa_temp_loss}
\end{equation}
where $\odot$ denotes the element-wise multiplication.
During inference, starting from a 
noisy latent initialized as 
white Gaussian noise, we can recover the noise-free latent code $\mathbf{z}_0$ through the reverse diffusion process. The recovered latent~$\mathbf{z}_0$ is then decoded using the pretrained VAE decoder~\cite{kingma2013auto}, yielding the inpainted image~$I_{t+1}^t$. 

In the following, we introduce the generation of the ground-truth cross-spatio-temporal image~$\hat{I}_{t+1}^t$ for network training.
To the best of our knowledge, no existing real-world SLAM dataset provides such images.
While a simulator-based strategy is plausible, the domain gap between synthetic and real-world scenes makes the simulated images unsuitable for our task.
To solve this problem, we adopt a 4D Gaussian splatting approach~\cite{chen2025omnire}. Given multiple real images~$\{ \cdots, I_{t-1}, I_t, I_{t+1}, \cdots \}$ obtained in a dynamic scene, we perform high-fidelity reconstruction of both static background and dynamic foreground. Then we render the reconstructed Gaussians at time~$t$ from the $(t+1)$-th view into a photo-realistic image~$\hat{I}_{t+1}^t$. We treat~$\hat{I}_{t+1}^t$ as the ground-truth cross-spatio-temporal image.

\subsection{Camera Localization} 
\label{subsec:cam_loc}
Our aim is to compute the pose~$\mathbf{T}_{t+1}$ of the $(t+1)$-th camera frame, given the estimate of the pose $\mathbf{T}_t$ of the $t$-th camera, and a scene representation at time $t$ encoded as a set of Gaussians $\bar{\mathcal{G}}_t$.\footnote{In our context, ``$\bar{\quad}$'' (e.g., $\bar{\mathcal{G}}_t$ and $\bar{\mathcal{P}}_t$) represents prior information, which is used to distinguish it from the estimated value.} 
We first estimate the relative transformation~$\Delta\mathbf{T}_{t \rightarrow t+1}$ from the $t$-th to the $(t+1)$-th camera frame leveraging photometric and geometric constraints, that we describe below. Once $\Delta \mathbf{T}_{t \rightarrow t+1}$ is obtained, ~$\mathbf{T}_{t+1}$ is obtained as $\mathbf{T}_{t+1} = \Delta \mathbf{T}_{t \rightarrow t+1} \mathbf{T}_t$.

\textbf{Photometric Constraints.}
Recall that, typically, photometric constraints 
are only applied to the static scene parts (see Fig.~\ref{fig:diff_camera}(a)). In contrast, leveraging our dreamed cross-spatio-temporal images, we introduce a novel photometric constraint that incorporates the dynamic foreground, thereby providing richer supervisory signals. As shown in Fig.~\ref{fig:diff_camera}(b), we transform the 
foreground and background Gaussians $\bar{\mathcal{G}}_t$ at time $t$ from the $t$-th camera frame to the $(t+1)$-th one using the transformation $\Delta\mathbf{T}_{t \rightarrow t+1}$. Then we render these Gaussians from the $(t+1)$-th view into an image:
\begin{equation}
\tilde{I}_{t+1}^{t}(\Delta\mathbf{T}_{t \rightarrow t+1})=\pi[\bar{\mathcal{G}}_t, \Delta\mathbf{T}_{t \rightarrow t+1}].
\label{eq:photometric_loss_pose}
\end{equation}
The rendered image
$\tilde{I}_{t+1}^{t}(\Delta\mathbf{T}_{t \rightarrow t+1})$ depends on the transformation~$\Delta\mathbf{T}_{t \rightarrow t+1}$.
Since~$\tilde{I}_{t+1}^{t}$ depicts both dynamic foreground and static background at time~$t$ from the $(t+1)$-th view, it corresponds to the above cross-spatio-temporal image $I_{t+1}^{t}$. We use the 
images $\tilde{I}_{t+1}^{t}$ and $I_{t+1}^{t}$
to formulate the photometric loss, optimizing the transformation $\Delta\mathbf{T}_{t \rightarrow t+1}$ as
\begin{equation}
        \min_{\Delta\mathbf{T}_{t \rightarrow t+1}} 
      \mathcal{L}_{\textnormal{photo}} \big( I_{t+1}^t, \tilde{I}_{t+1}^t( \Delta\mathbf{T}_{t \rightarrow t+1})\big).
      \label{eq:pose_photo_loss}
\end{equation}
By jointly leveraging the information of both static and dynamic content, our approach significantly enhances localization accuracy, as validated through experiments. 
In practice, to conduct optimization in Eq.~(\ref{eq:pose_photo_loss}), we need a relatively reliable initial value of the transformation~$\Delta\mathbf{T}_{t \rightarrow t+1}$. We obtain it based on the following geometric constraint.

\textbf{Geometric Constraints.} 
We feed the images $I_{t+1}$ and $I_t$ to our Gaussian prediction network introduced below.
As partial output, we obtain a set of Gaussians $\mathcal{G}_{t}$, which corresponds to the 3D scene at time~$t$, in the $(t+1)$-th camera frame. 
Then we extract the centers of $\mathcal{G}_{t}$ and the above prior Gaussians~$\bar{\mathcal{G}}_{t}$ 
 respectively, obtaining
point clouds 
$\mathcal{P}_{t}$ in the $(t+1)$-th camera frame and $\bar{\mathcal{P}}_{t}$ in the $t$-th camera frame. 
Since $\mathcal{P}_{t}$ and $\tilde{\mathcal{P}}_{t}$
are associated with the same image~$I_t$, each pair of points is inherently associated through their corresponding pixel.
Accordingly, we can establish a set of point correspondences~$\{(\mathbf{p}_{k}, \bar{\mathbf{p}}_{k})\}$ between $\mathcal{P}_{t}$ and $\bar{\mathcal{P}}_{t}$. Given these point correspondences, we formulate the following point cloud alignment loss based on the transformation $\Delta \mathbf{T}_{t \rightarrow t+1}$:
\begin{equation}
    \min_{\Delta\mathbf{T}_{t \rightarrow t+1}} \sum_k \|  \mathbf{p}_{k} -  \Delta\mathbf{T}_{t \rightarrow t+1} (\bar{\mathbf{p}}_{k} )  \|_2^2 \ .
\end{equation}
Specifically, $\Delta \mathbf{T}_{t \rightarrow t+1}$ is obtained via singular value decomposition (SVD)~\cite{icp_pami}, and used as the initial seed for the photometric constraint-based optimization in Eq.~(\ref{eq:pose_photo_loss}). Compared with the existing methods targeting static environments~\cite{mast3rslam}, our strategy has two strengths. First, it leverages the dynamic foreground to better compensate for noise. Second, it leads to a higher degree of overlap between point clouds for the alignment, providing a higher number of constraints.

\subsection{3D Scene Mapping}
\label{sec:Geometry_model}
We propose a novel network
that
can predict both foreground and background Gaussians in a feedforward manner. It can significantly improve the efficiency while maintaining high accuracy, compared with classical Gaussian splatting-based SLAM~\cite{PG-SLAM, WildGS}. Moreover, we leverage the dreamed cross-spatio-temporal images to further refine the Gaussians.

\begin{figure}[!t]
\footnotesize
\centering
\renewcommand{\tabcolsep}{0.3pt}
\begin{tabular}{cc}
\includegraphics[width=0.95\linewidth]{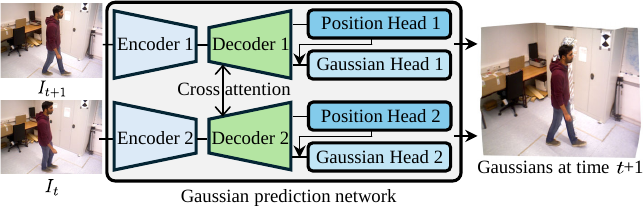}
\\
(a)
\\
\includegraphics[height=0.5\linewidth]{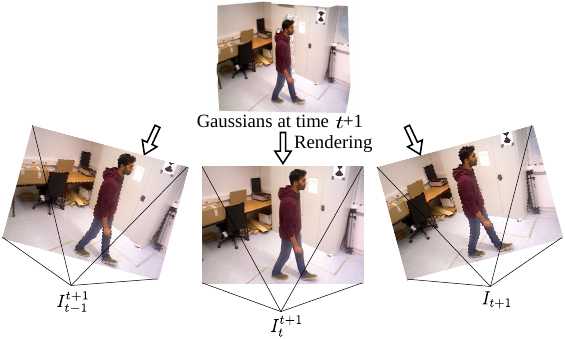}
\\
(b)
\end{tabular}
\caption{\textbf{3D Gaussian prediction and refinement.}
 (a) Given images~$I_{t+1}$ and $I_t$, we design a feedforward network to predict dynamic Gaussians at both time $t+1$ and time $t$. Here, we only visualize the predicted Gaussians at time~$t+1$. (b) We refine Gaussians at time $t+1$ based on the photometric loss regarding both dreamed cross-spatio-temporal images~$I_{t-1}^{t+1}, I_{t}^{t+1}$ and real image~$I_{t+1}$. These images depict the same scene content at time $t+1$ from the $(t-1)$-th, $t$-th, and $(t+1)$-th views, respectively.}
\label{fig:mapping}
\end{figure}

\textbf{Feedforward Gaussian Prediction.}
Fig.~\ref{fig:mapping}(a) shows the architecture of our network, which consists of two modules. The first module regresses Gaussian positions, similarly to the foundation model introduced in Section~\ref{sec:Preliminaries}.
In brief,
given images $I_{t+1}$ and $I_t$, we obtain two sets of image tokens $\{G^i_1\}_{i=1}^N$ and $\{G^i_2\}_{i=1}^N$.
These tokens are fed to two position heads, outputting the Gaussian centers (point clouds), both in the $(t+1)$-th camera frame. The second module, which is our main contribution, predicts the rest of the attributes for each Gaussian, using the Gaussian centers predicted by the first module as geometric guidance.
Specifically, in this second module, we first leverage the Point Transformer V3~\cite{wu2024PointTrans} in order to extract point cloud features.
We then use $N$ zero-convolution layers to map these features into point tokens $\{K^i_1\}_{i=1}^N$ and $\{K^i_2\}_{i=1}^N$.
This mapping does not involve cross attention-based blocks,  
since the integration of this information has been completed before generating the point clouds.
Further, taking the first branch as an example, we fuse the above point and image tokens by
$\hat{G}_1^i = G_1^i + K_1^i$. 
Finally, we introduce Gaussian heads
$\mathcal{H}_1$ and $\mathcal{H}_2$
to respectively predict Gaussian attributes~$\mathcal{G}_{t+1}$ and $\mathcal{G}_{t}$,
along with their associated confidences~$\mathcal{C}_{t+1}$ and~$\mathcal{C}_{t}$:
\begin{subequations}
    \begin{align}
       \mathcal{G}_{t+1}, \mathcal{C}_{t+1} &= \mathcal{H}_1(\hat{G}^0_1, \ldots, \hat{G}_1^N), \label{eq:recon_a} \\
    \mathcal{G}_t, \mathcal{C}_t &= \mathcal{H}_2(\hat{G}_2^0, \ldots, \hat{G}_2^N). \label{eq:recon_b}
\end{align}
\end{subequations}
As the map grows, per-pixel Gaussian predictions may cause redundancy. In that case, we prune Gaussians based on their local density to minimize storage overhead. Notably, pruning is applied only to Gaussians that are no longer tracked, to avoid performance degradation.

We train our Gaussian prediction network~$\mathcal{R}$ as follows.
Firstly, we use the photometric loss to enforce the
appearance constraint. Without loss of generality, we take the predicted Gaussians~$\mathcal{G}_{t+1}(\mathcal{R})$ and their associated  confidences~$\mathcal{C}_{t+1}(\mathcal{R})$ in Eq.~(\ref{eq:recon_a}) as an illustrative example.
From the $(t+1)$-th view, we render these Gaussians 
into an image~$\tilde{I}_{t+1}$, which innovatively incorporates the confidences as weights: 
\begin{equation}\tilde{I}_{t+1}(\mathcal{R}) = \pi[\mathcal{G}_{t+1}(\mathcal{R}) \cdot \mathcal{C}_{t+1}(\mathcal{R})].\end{equation} We then optimize the weights of the network $\mathcal{R}$ by minimizing the difference between the rendered image $\tilde{I}_{t+1}(\mathcal{R})$ and the ground-truth image $I_{t+1}$ to optimize the network $\mathcal{R}$, specifically 
$
        \min_{\mathcal{R}} \mathcal{L}_{\textnormal{photo}}\big( I_{t+1},\tilde{I}_{t+1}(\mathcal{R})\big)$.
In addition, we extract point clouds from Gaussians, and apply the geometric loss~\cite{zhang2024monst3r} for network training.

\textbf{Gaussian Refinement.}
The predicted Gaussians 
should be reliable. However, since the dynamic foreground at a given time is constrained by only a single view, there remains room for improving accuracy.
For this reason, we refine the Gaussians based not only on the real images, but also the dreamed cross-spatio-temporal images. This refinement can use different numbers of dreamed images. We empirically observed that two dreamed images achieve the best balance between accuracy and efficiency, and thus we introduce this case in the following.

As shown in Fig.~\ref{fig:mapping}(b), our aim is to optimize the foreground and background Gaussians $\mathcal{G}_{t+1}$ at time~$t+1$ using multi-view constraints. Following Section~\ref{Spatio-temporal}, we first use the real image pairs~$(I_{t+1}, I_t)$ and $(I_{t+1}, I_{t-1})$ to dream the cross-spatio-temporal images~$I_{t}^{t+1}$ and~$I_{t-1}^{t+1}$, respectively.
These images depict the Gaussians $\mathcal{G}_{t+1}$ from the $t$-th and $(t-1)$-th views, respectively.
Then, with the poses of the $(t+1)$-th and $t$-th cameras (obtained as described in Section~\ref{subsec:cam_loc}), we transform the Gaussians $\mathcal{G}_{t+1}$ from the $(t+1)$-th camera frame to the $t$-th camera frame, and further render them into an image $\tilde{I}_{t}^{t+1}(\mathcal{G}_{t+1})$. Similarly, 
we also transform the Gaussians $\mathcal{G}_{t+1}$ 
to the $(t-1)$-th camera frame, and further render them into an image $\tilde{I}_{t-1}^{t+1}(\mathcal{G}_{t+1})$. These rendered images respectively correspond to the above cross-spatio-temporal images~$I_{t}^{t+1}$ and~$I_{t-1}^{t+1}$, which can provide effective supervisory constraints on Gaussians $\mathcal{G}_{t+1}$. We formulate such constraints with the following photometric loss:
\begin{equation}
    \min_{\mathcal{G}_{t+1}}
    \mathcal{L}_{\textnormal{photo}}\big(I_{t}^{t+1}, \tilde{I}_{t}^{t+1}(\mathcal{G}_{t+1} )\big)
    +
    \mathcal{L}_{\textnormal{photo}}\big(I_{t-1}^{t+1}, \tilde{I}_{t-1}^{t+1}(\mathcal{G}_{t+1} )\big).
    \label{eq:gau_optim_loss}
\end{equation}
In addition, we also formulate complementary constraints leveraging the real images.
Specifically, we render Gaussians~$\mathcal{G}_{t+1}$ from the $(t+1)$-th view into an image~$\tilde{I}_{t+1}(\mathcal{G}_{t+1})$, which correspond to the real image~$I_{t+1}$, and we define the photometric loss as $\min_{\mathcal{G}_{t+1}} \mathcal{L}_{\textnormal{photo}}\big(I_{t+1}, \tilde{I}_{t+1}(\mathcal{G}_{t+1} )\big)$. 

\section{Exploration Planning} 
\label{sec:planning}

In this section, we present how we plan a farsighted
exploration path. Our main technical contribution is to dream semantically plausible structures of unexplored regions, and then leverage these structures to make decisions. Such a strategy can significantly reduce the path length and improve the exploration completeness. To facilitate understanding, we first briefly introduce the pipeline, and then highlight our contributions.

\subsection{Exploration Planning Pipeline}

We formulate the exploration as the online generation of a set of candidate waypoints, and the selection of the optimal one to move to.

\textbf{Generation of Waypoints.}
We make use of our 3D scene mapping method, introduced above, to reconstruct the scene with a 3D Gaussian map. To extract traversable areas in the 3D space, we render Gaussians into a top-view opacity image, which is then binarized as a free-space map. We then use Voronoi diagrams~\cite{okabe2009spatial} to convert this free-space map into a 2D topological map with multiple nodes and edges, the former corresponding to waypoints and the latter representing their connections. As the Gaussian map grows, we update the topological map with newly added nodes.

\textbf{Motion to Optimal Waypoints.}
We first cluster waypoints according to their distances in the topological map. Accordingly, the environment is partitioned into several sub-regions,
each of which is associated with a representative waypoint. We perform a hierarchical planning by globally ordering all sub-regions and locally exploring each of them. In terms of global ordering, we determine the shortest path to sequentially connect all the representative waypoints of the sub-regions. 
We cast this planning as a solvable traveling salesman problem~\cite{razali2011genetic}. Then the robot moves to the first sub-region in the planned sequence, followed by re-performing the global ordering on both remaining and added sub-regions. The added sub-regions include both newly observed and dreamed sub-regions (sub-region dreaming is our main technical novelty).
The reason for re-planning is that the added sub-regions may overturn the optimality of the previous path. In terms of local planning within a sub-region, the robot preferentially moves to the waypoints where the surroundings are under-explored. This strategy targets a quick gain of information on the map. The traversable path between waypoints is computed by~\cite{dijkstra2022note}.

\subsection{Structure Dreaming-based Global Ordering} 
\label{sec:plan_gobal_Predict}

When performing the global ordering of sub-regions, 
a common strategy is to leverage the observed structures. However, in this case, many structures are only partially observed, and thus the information is incomplete, resulting in unnecessary detours. To solve this problem, we propose to dream semantically plausible structures of the unobserved areas.
\begin{figure}[!t]
\footnotesize
\centering
\includegraphics[width=1.0\linewidth]{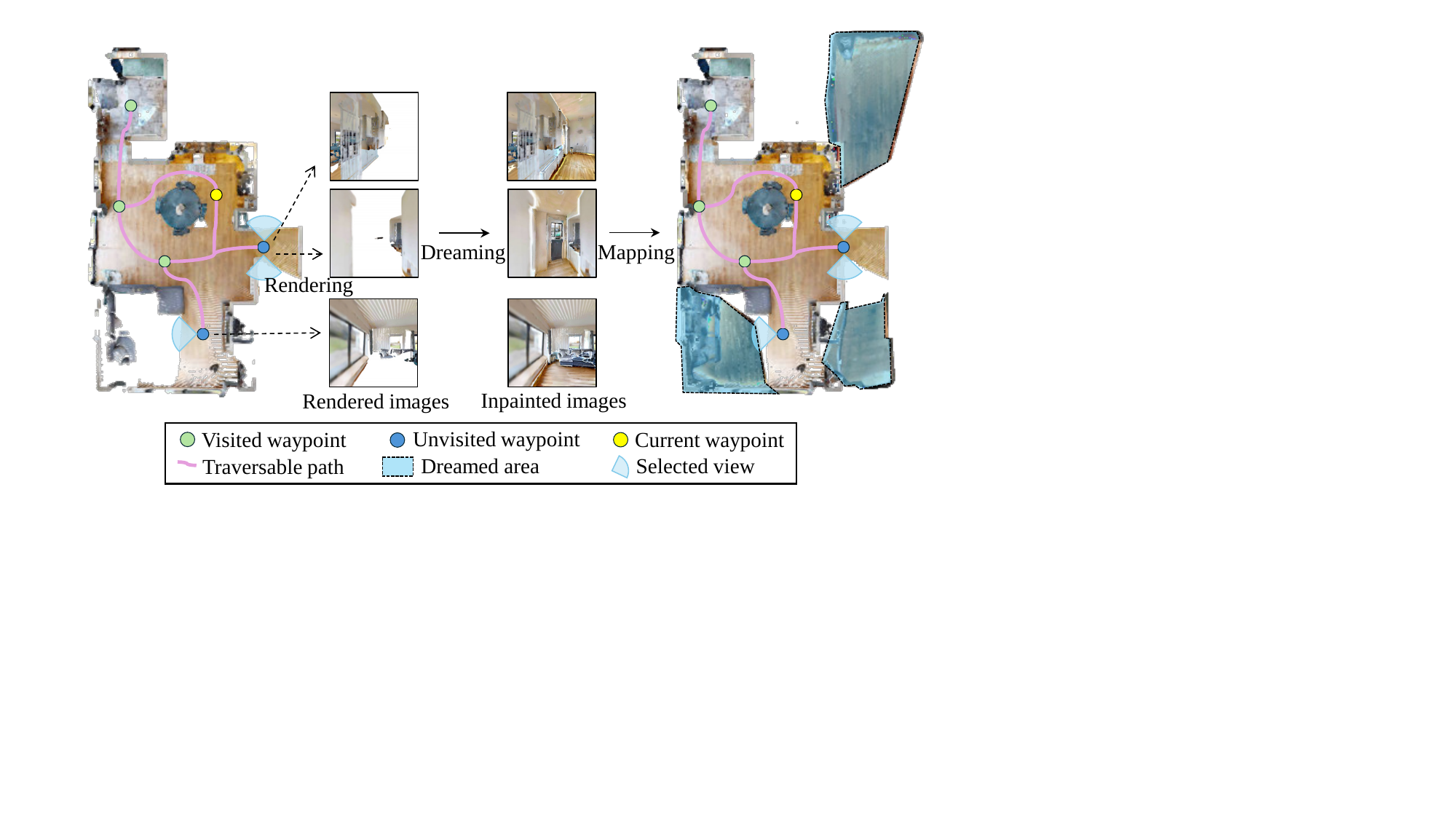}
\caption{\textbf{Dreaming semantically plausible structures of unexplored areas.}
At an unvisited waypoint, we place virtual cameras to render images from different views and select the suitable images. Then we inpaint the selected images, and use them to predict Gaussians. By integrating the dreamed Gaussians into the existing Gaussians, we obtain more complete structures of the environment.}
\label{fig:dream_scene}
\end{figure}

\textbf{Dreaming Semantically Plausible Structures.}
As shown in Fig.~\ref{fig:dream_scene}, at each unvisited waypoint within a subregion, we first render the reconstructed Gaussians into opacity~$\{\tilde{O}_i\}_{i=1}^4$ and RGB images~$\{\tilde{I}_i\}_{i=1}^4$ from four non-overlapping views (each of them with a field of view of $90^\circ$). These views collectively cover the full panorama around the waypoint. Intuitively, if a pixel of the image~$\tilde{O}_i$ has a high opacity, it is occupied and corresponds to a mapped structure. For each image~$\tilde{O}_i$ or $\tilde{I}_i$ with $N$ pixels, we compute its occupancy score~$g_i$ defined as the proportion of 
occupied pixels:
\begin{equation}
    g_i = \sum_{j=1}^N \mathbb{I} \{ \tilde{o}_{j} > \tau_\textnormal{o} \} / N,
     \label{eq:score}
\end{equation}
where $\tilde{o}_{j}$ denotes the opacity of the $j$-th pixel of the image~$\tilde{O}_{i}$, $\tau_\textnormal{o}$ denotes the opacity threshold, and $\mathbb{I} \{ \cdot \}$ returns 1 if the condition is satisfied. 
As to the suitability of the image~$\tilde{I}_{i}$ for structure dreaming, we disregard both too high and too low occupancy scores~$g_i$. Excessively high scores mean that the environment has been well-observed and is unnecessary to dream, while excessively low scores mean that the observations are insufficient to provide reliable semantic references for dreaming. Therefore, we only consider view(s) whose score~$g_i$ falls within a range $[\underline{\tau}, \overline{\tau}]$ for inpainting. We inpaint the images~$\{ \tilde{I}_i \}$ associated with the selected views in a semantically plausible way (details will be introduced in Section~\ref{sec:path_dreanm}).

\begin{figure}[!t]
\footnotesize
\centering
\begin{tabular}{c}
\includegraphics[width=0.95\linewidth]{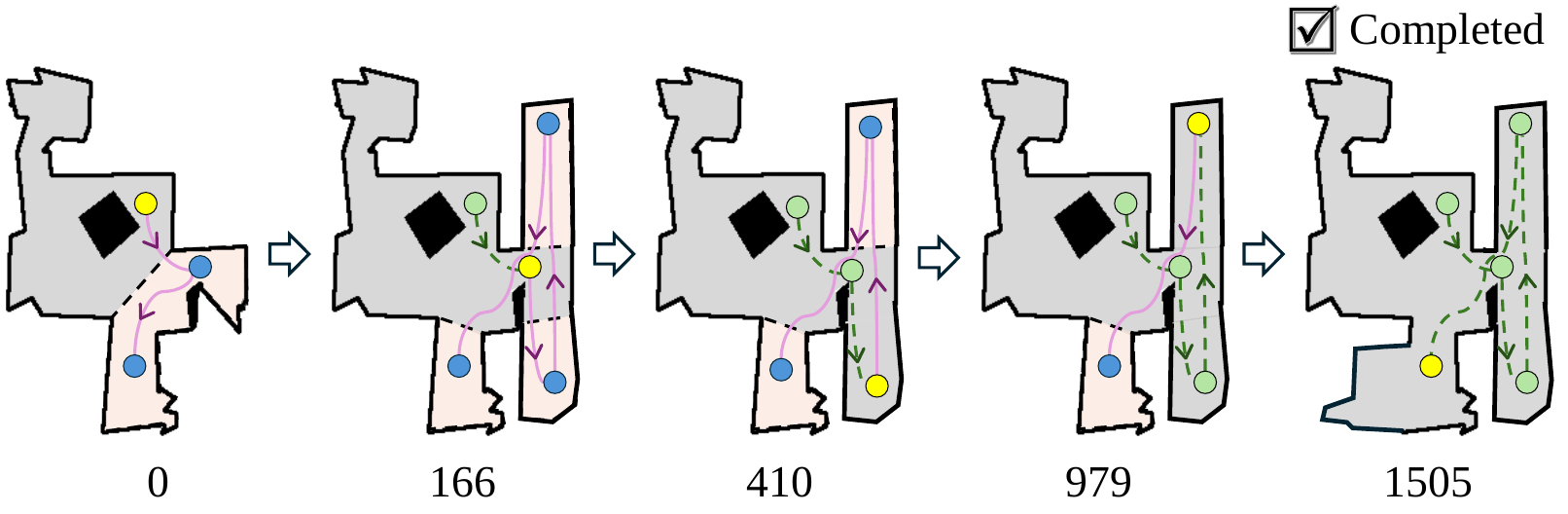} \\ 
(a) \\ 
\includegraphics[width=0.95\linewidth]{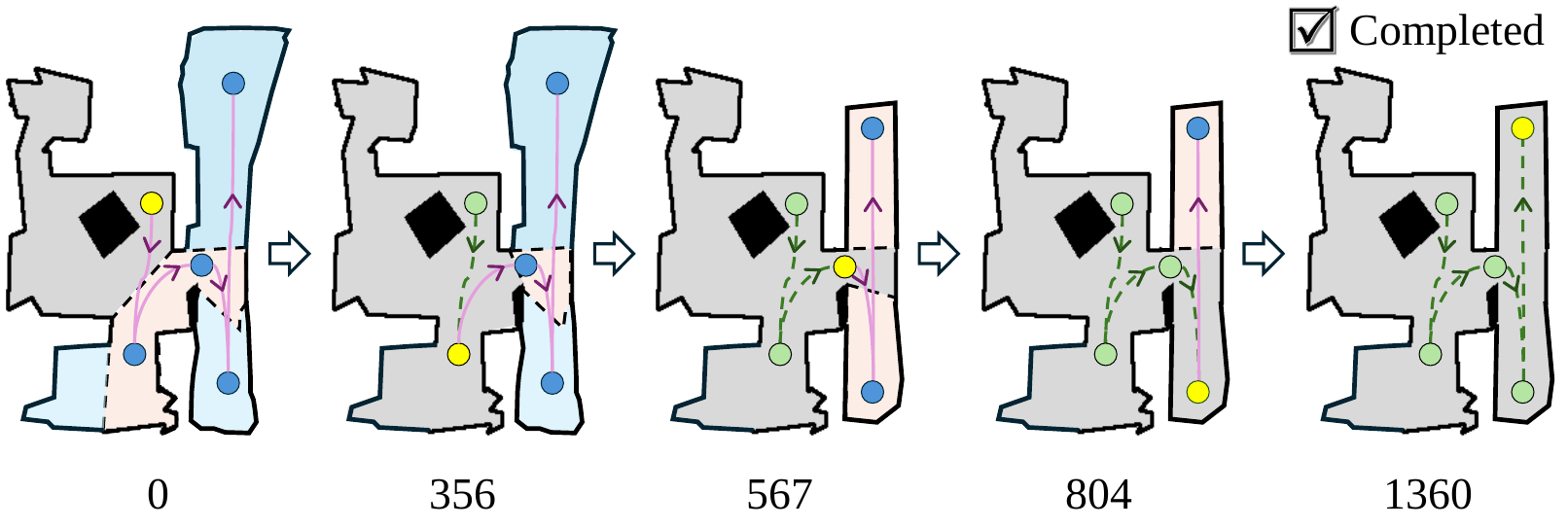} \\
(b) \\
\includegraphics[width=0.8\linewidth]{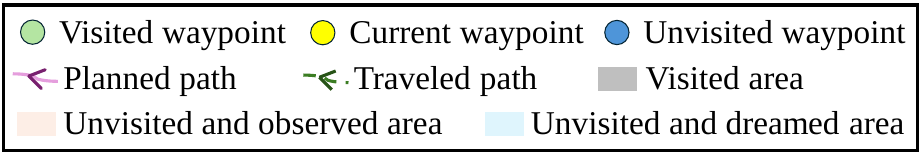}
\end{tabular}

\caption{\textbf{Comparison between global ordering strategies without and with the capability of structure dreaming.}
In this schematic diagram, the number below each image represents the accumulated length of the traveled path. They do not have a specific unit, but show relative magnitudes of values.
(a) Strategies without dreaming capability 
result in suboptimal trajectories with unnecessary detours. (b) Our method leverages both observed and dreamed structures to plan a farsighted path.
}
\label{fig:planning}
\end{figure}

\textbf{Global Ordering.}
Given a set of inpainted images, we first predict new Gaussians using the Gaussian prediction network that we described before. Benefiting from high-quality images, these Gaussians are semantically plausible to describe the underlying structures of the unobserved areas. After that, we integrate the dreamed Gaussians into the existing Gaussian map. Such an enriched map leads to a refined topological map and additional sub-regions. As shown in Fig.~\ref{fig:planning}, especially for long-horizon planning, our approach using the dreamed structures significantly outperforms the shortsighted planning method without the dreaming capability. Our method globally shortens the path length and yields higher mapping efficiency. Please note that while our dreamed structures are semantically plausible and effective for planning, they inevitably differ from the real structures.
Accordingly, for the photo-realistic scene mapping, we differentiate between the dreamed and observed sub-regions during exploration. After the robot visits a dreamed sub-region, it uses real observations to update the representation of this sub-region, replacing the previously dreamed structures.

\textbf{Dynamic Environments.}
Our planning method can handle dynamic environments well. This is mainly attributed to our reconstructed map, in which we can differentiate between the static background and dynamic foreground. First, existing methods~\cite{ActiveSplat,kuang2024active} mistakenly reconstruct the foreground as the background, which causes structure blur and further blocks the feasible traversable path. By contrast, our method can distinguish and only use the structures of static background for path planning, avoiding the occlusion of the paths to unexplored sub-regions. Please refer to the supplementary material for details.
Second, during the structure dreaming, we only render the static background into images on purpose. This operation contributes to a holistic scene layout prediction without being affected by the foreground occlusion.
Third, dynamic objects are typical obstacles for robot movement. Our 
mapping approach can determine the shapes and relative locations of these dynamic obstacles at each time. By feeding this information to Dijkstra's algorithm~\cite{dijkstra2022note}, the robot can effectively avoid obstacles.

\subsection{Semantically Plausible Image Inpainting}
\label{sec:path_dreanm}
Recall that in the above subsection, we inpaint the rendered images~$\{\tilde{I}_i\}$. We achieve this by proposing a network that can effectively leverage the surrounding information to achieve a semantically plausible inpainting. It is based on the diffusion model, which is similar to our inpainting network in Section~\ref{Spatio-temporal}. Their main differences lie in the inpainting mask definition and reference code generation.

For one thing, we define an inpainting mask $\mathbf{B}$ to indicate the pixels to inpaint. During training, we randomly generate~$\mathbf{B}$ of the complete image~$\hat{I}_i$ to improve the generalization of the network. At the inference stage, the inpainted pixels of the rendered image~$\tilde{I}_i$ are defined as a set of non-occupied pixels whose opacity is lower than the threshold $\tau_{\textnormal{o}}$ (see Eq.~(\ref{eq:score})).
For another, we introduce how we generate the reference code as follows. 
We use the complementary mask of the above inpainting mask $\mathbf{B}$ to indicate the occupied pixels. During training, given the masked image~$\hat{I}_i$, 
we encode its occupied pixels into the reference code~$\mathbf{c}$ based on the pretrained variational autoencoder~\cite{kingma2013auto}.
At inference time, we use this encoder to encode the occupied pixels of the rendered image~$\tilde{I}_i$ 
as the reference code~$\mathbf{c}$. The code~$\mathbf{c}$ can provide the structural and appearance information of the surrounding pixels for reliable inpainting.

Given the above inpainting mask~$\mathbf{B}$ and reference code~$\mathbf{c}$, we introduce the diffusion-based inpainting.
In the forward process, we generate the noisy code $\mathbf{z}_s$ at step~$s$. In the reverse process, we introduce a network $\mathcal{P}$ to predict the added noise~$\epsilon$. This network incorporates the code~$\mathbf{c}$ and the downsampled mask~$\bar{\mathbf{B}}$ compatible with the noise dimension:
\begin{equation}
   \epsilon = \mathcal{P}(\mathbf{z}_s, s, \mathbf{c}, \bar{\mathbf{B}}). 
   \label{eq:error_planning}
\end{equation}
To train the network~$\mathcal{P}$, we 
use the following loss:
\begin{equation}
    \mathcal{L}=
    \mathbb{E} \Big[\| \bar{\mathbf{B}} \odot \big(\hat{\epsilon} - \epsilon(\mathcal{P}) \big)\|_2 \Big].
\end{equation}
During inference, 
given a noisy code expressed by white noise, we follow the reverse process to obtain the denoised code~$\mathbf{z}_0$. By decoding~$\mathbf{z}_0$ based on the pretrained decoder~\cite{kingma2013auto}, we generate the inpainted image. This image exhibits a semantically plausible appearance where the inpainted areas are harmonious with the observed areas.

\begin{table*}[!t]
	\renewcommand\arraystretch{1.3} 
	\centering
	\caption{Camera localization comparisons 
    on the \textsf{TUM}~\cite{tumbenchmark} and \textsf{Bonn}~\cite{ReFusion} datasets. 
	}
	\begin{threeparttable}
		\begin{tabular}{c}
        \renewcommand{\tabcolsep}{2.86pt}
			\renewcommand\arraystretch{1.3}
			\begin{tabular}{c|cc|cc|cc|cc|cc|cc|cc||cc}
            \toprule
				\multicolumn{17}{c}{\textsf{TUM} dataset~\cite{tumbenchmark}} \\
				\midrule
				
				\multicolumn{1}{c}{Sequences} & \multicolumn{2}{c}{\texttt{f3/wk\_xyz}} & \multicolumn{2}{c}{\texttt{f3/wk\_hf}} & \multicolumn{2}{c}{\texttt{f3/wk\_st}} & \multicolumn{2}{c}{\texttt{f3/st\_hf}}   & \multicolumn{2}{c}{\texttt{f3/st\_rpy}}  & \multicolumn{2}{c}{\texttt{f3/st\_st}}  & \multicolumn{2}{c}{\texttt{f3/st\_xyz}} & \multicolumn{2}{c}{Average} \\
				\midrule
				\textit{}  & RMSE$\downarrow$ & SD$\downarrow$ & RMSE$\downarrow$ & SD$\downarrow$  &  RMSE$\downarrow$ & SD$\downarrow$   & RMSE$\downarrow$ & SD$\downarrow$ & RMSE$\downarrow$ & SD$\downarrow$ & RMSE$\downarrow$ & SD$\downarrow$ & RMSE$\downarrow$ & SD$\downarrow$ & RMSE$\downarrow$ & SD$\downarrow$\\
                \textsf{ORB-SLAM3}~\cite{Campos2021orb} &28.1 & 12.2 & 30.5 & 9.0 & 2.0 & 1.1  & 2.6 & 1.6  & 6.4 & 2.5 & 0.98 & 0.46 & 1.6 & 0.7 & 10.31  & 3.94  \\

                \textsf{MonST3R}~\cite{zhang2024monst3r} & 26.7 & 13.2 & 44.6 & 23.9 & 1.7  & 0.8 & 38.7 & 20.2 & 5.5 & 2.6 & 2.2 & 1.2 & 30.2 & 13.2 & 21.37 & 10.73 \\
                
				\textsf{RoDyn-SLAM}~\cite{jiang2024rodyn}  & 8.3 & 5.5 & 5.6 & 2.8 & 1.7  & 0.9 & 4.4 & 2.2 & 11.4 & 4.6 & 0.76 & 0.43 & 5.0 & 1.0 & 5.31 & 2.49 \\

                \textsf{PG-SLAM}~\cite{PG-SLAM} & 6.8 & 2.9 & 11.7 & 4.4 & 1.4  & 0.6 & 4.0 & 1.5 & 5.4 & 2.4 & 0.72 & 0.39 & 1.5 & 0.5 & 4.50 & 1.81\\
                
				\textsf{WildGS-SLAM}~\cite{WildGS} & 
                \textbf{1.3} & 
                \textbf{0.6} &  
                \textbf{1.6} &  
                {0.8} & 
                {0.4} & 
                {0.2} & 
                {2.0} & 
                \textbf{0.9} & 
                {2.4} & 
                {1.4} & 
                {0.5} & 
                {0.3} & 
                {0.8} & 
                {0.4} & 
                {1.28} & 
                {0.65} 
                \\

				\textsf{Dream-SLAM (ours)}
                & {1.7} & 
                {0.7} &
                \textbf{1.6} &
                \textbf{0.7} &
                \textbf{0.3} &
                \textbf{0.1} &
                \textbf{1.9} &
                \textbf{0.9} &
                \textbf{2.3} &
                \textbf{0.7} &
                \textbf{0.3} &
                \textbf{0.1} &
                \textbf{0.6} &
                \textbf{0.2} &
                \textbf{1.27} &
                \textbf{0.48} \\
				\midrule
				\midrule
			\end{tabular}
            \\
			\renewcommand{\tabcolsep}{4.6pt}
			\renewcommand\arraystretch{1.3}
			\begin{tabular}{c|cc|cc|cc|cc|cc|cc||cc}
				
				\multicolumn{15}{c}{\textsf{Bonn} dataset~\cite{ReFusion}}\\
				\midrule
				\multicolumn{1}{c}{Sequences}  & \multicolumn{2}{c}{\texttt{balloon}} & \multicolumn{2}{c}{\texttt{balloon2}} & \multicolumn{2}{c}{\texttt{ps\_track}} & \multicolumn{2}{c}{\texttt{ps\_track2}} & \multicolumn{2}{c}{\texttt{mv\_box}} & \multicolumn{2}{c}{\texttt{mv\_box2}}  & \multicolumn{2}{c}{Average} \\
				\midrule
				
				\textit{}  & RMSE$\downarrow$ & SD$\downarrow$  & RMSE$\downarrow$ & SD$\downarrow$  & RMSE$\downarrow$ & SD$\downarrow$  &  RMSE$\downarrow$ & SD$\downarrow$ &  RMSE$\downarrow$ & SD$\downarrow$   & RMSE$\downarrow$ & SD$\downarrow$   & RMSE$\downarrow$ & SD$\downarrow$\\

                \textsf{ORB-SLAM3}~\cite{Campos2021orb} & 6.5 & 2.9  & 17.7 & 8.6 &70.7 & 32.6  & 77.9 & 43.8  & 28.0 & 8.4 & 3.5 & 1.5 & 34.05 & 16.30\\

                \textsf{MonST3R}~\cite{zhang2024monst3r} & 11.1 & 8.5 &  14.3 & 6.3 & 25.4 & 15.8 & 22.5 & 9.4 &  7.6 & 3.5 & 13.1 & 5.3 & 15.67 & 8.13\\

				\textsf{RoDyn-SLAM}~\cite{jiang2024rodyn}  & 7.9 & 2.7 & 11.5 & 6.1  & 14.5  & 4.6 & 13.8 & 3.5  & 7.2 & 2.4 & 12.6  & 4.7 & 11.25 & 4.00\\

				\textsf{PG-SLAM}~\cite{PG-SLAM}  & 6.4 & 2.2 & 7.3 & 3.4  & 5.0  & 1.9 & 8.5 & 2.8  & 4.6 & 1.3 & 7.0  & 2.0 & 6.47 & 2.27\\

                \textsf{WildGS-SLAM}~\cite{WildGS} &
                {2.8} &
                {1.2} &
                {2.4} &
                {1.1} &
                {3.1} &
                {2.0} &
                {3.0} &
                {1.3} &
                {1.6} &
                {0.8} &
                {2.2} &
                {1.2} &
                {2.52} &
                {1.27}
                \\
                
				  \textsf{Dream-SLAM (ours)} &
                \textbf{1.9} &
                \textbf{0.7} &
                \textbf{1.9} &
                \textbf{0.4} &
                \textbf{1.5} &
                \textbf{0.5} &
                \textbf{2.7} &
                \textbf{0.8} &
                \textbf{0.6} &
                \textbf{0.2} &
                \textbf{1.4} &
                \textbf{0.5} &
                \textbf{1.67} &
                \textbf{0.52}\\

				\bottomrule 
			\end{tabular}

	\end{tabular}
	
\end{threeparttable}

\label{table:com_traj}

\end{table*}
\begin{figure}[!t]
\footnotesize
	\renewcommand\arraystretch{1.2}
	\renewcommand{\tabcolsep}{1pt} 
	\centering
\begin{tabular}{cc}
 \multicolumn{1}{c}{\includegraphics[width=0.5\linewidth]{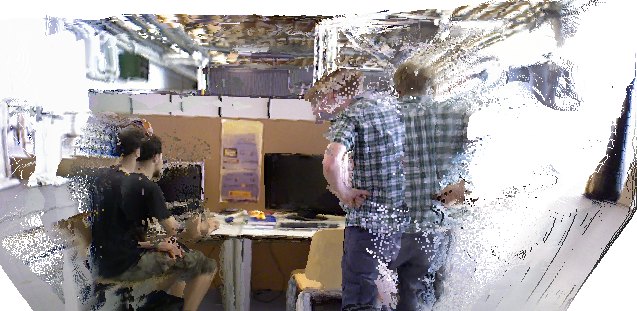}}  & \multicolumn{1}{c}{\includegraphics[width=0.42\linewidth]{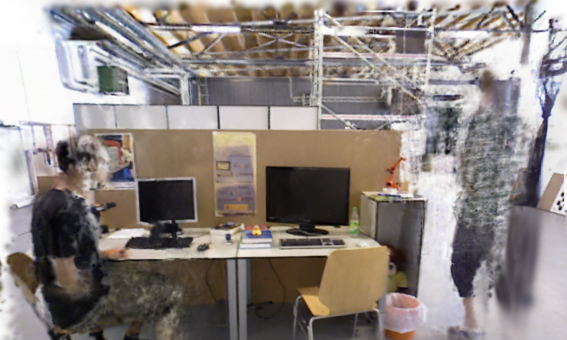}}  \\
(a) \textsf{MonST3R}~\cite{zhang2024monst3r} &
(b) \textsf{WildGS-SLAM}~\cite{WildGS}  \\[0.5em]
\multicolumn{2}{c}{\includegraphics[width=0.75\linewidth]{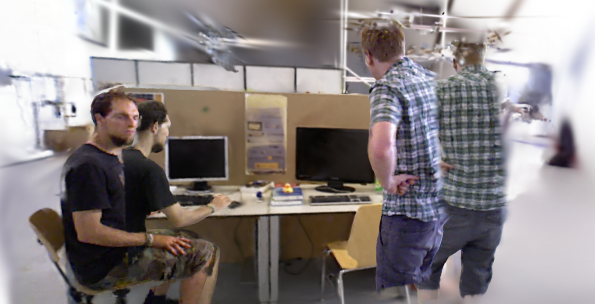}} \\
\multicolumn{2}{c}{(c) \textsf{PG-SLAM}~\cite{PG-SLAM} } \\[0.5em]
\multicolumn{2}{c}{\includegraphics[width=0.78\linewidth]{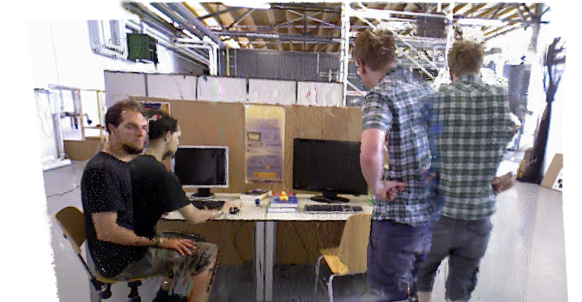}} \\
\multicolumn{2}{c}{(d) \textsf{Dream-SLAM} (ours) }
\end{tabular}
\caption{\textbf{Qualitative comparison 
of dynamic scene mapping methods on Sequence \texttt{wk\_xyz} in the \textsf{TUM} dataset~\cite{tumbenchmark}.} We render the mapped scenes from novel views. (a) \textsf{MonST3R}~\cite{zhang2024monst3r} does not support a photo-realistic rendering due to the point cloud representation. (b) \textsf{WildGS-SLAM}~\cite{WildGS} fails to map the dynamic foreground.
(c) \textsf{PG-SLAM}~\cite{PG-SLAM} can reconstruct both static background and  dynamic foreground, but some rendering regions remain low-quality.
(d) Our \textsf{Dream-SLAM} can map both background and foreground, and also achieves the highest mapping quality. }
       	\label{Tum_mapping}
\end{figure}

\section{Experiments on Public Datasets}
\label{exp_real}

To the best of our knowledge, currently, there is no public dataset that can simultaneously evaluate 1) the accuracy of localization and mapping, together with 2) the effectiveness of exploration planning. Therefore, in this section, we separately evaluate these two tasks. All the experiments are conducted on a server with an E3-1226 CPU and RTX 4090 GPU.
For a joint evaluation on our self-collected data, please refer to the next section.

\subsection{Localization and Mapping}

\subsubsection{Experimental Setup}
We introduce datasets, evaluation metrics, implementation details, and methods for comparison.

\textbf{Datasets.}
TUM RGB-D~\cite{tumbenchmark} (denoted by \textsf{TUM}) and Bonn RGB-D dynamic~\cite{ReFusion}  (denoted by \textsf{Bonn}) datasets are two well-known datasets to benchmark SLAM methods, especially in dynamic environments:
\begin{itemize}
\item The \textsf{TUM} dataset was collected in indoor environments. It contains multiple sequences with one or more people walking around. It provides ground-truth camera trajectories.
\item  
Compared with the \textsf{TUM} dataset, the \textsf{Bonn} dataset additionally includes scenarios involving human–object interactions, such as box carrying. Moreover, human movement on partial sequences is larger.
\end{itemize}

\textbf{Evaluation Metrics.}
To assess camera localization accuracy, we adopt the widely used absolute trajectory error~\cite{UZH_ATE}, which measures the difference between the estimated 
and the ground-truth trajectories. We present the error in terms of root mean square (RMSE)~\cite{tumbenchmark} and standard deviation (SD), both expressed in units of centimeters.
Regarding mapping quality, we evaluate the rendering results. For quantitative evaluation, we render the reconstructed scene from novel viewpoints. For qualitative analysis, we employ widely adopted metrics PSNR~\cite{hore2010image}, SSIM~\cite{ssim}, and LPIPS~\cite{zhang2018unreasonable} to measure the differences between the rendered and the ground-truth images. 

\textbf{Implementation Details.} 
For the diffusion model introduced in Section~\ref{Spatio-temporal}, we adopt a large-scale pretrained text-to-image 
network~\cite{rombach2022high} and apply LoRA-based fine-tuning with a rank of 8. For our Gaussian prediction network introduced in Section~\ref{sec:Geometry_model}, we initialize it with the pretrained MonST3R~\cite{zhang2024monst3r} and adopt a two-stage training strategy.
In the first stage, we fine-tune the position head and decoder based on the point loss, while in the second stage, we fix the other modules and train the Gaussian head only. As to the data for the above fine-tuning, we consider the \textsf{Neuman}~\cite{neuman} and \textsf{WildGS}~\cite{WildGS} datasets, which contain scenes with multiple people and moving objects.
This practice is helpful to evaluate the generalization when testing on the \textsf{TUM} and \textsf{Bonn} datasets.

\textbf{Methods for Comparison.} We compare the localization and mapping modules of our \textsf{Dream-SLAM} against the following state-of-the-art SLAM methods introduced in Section~\ref{sec:rel_work}:
\begin{itemize}
    \item \textsf{ORB-SLAM3}~\cite{Campos2021orb}: 
    A classic feature-based method. It leverages the epipolar geometry constraints to filter out dynamic objects.
    \item \textsf{Rodyn-SLAM}~\cite{jiang2024rodyn}: 
    An implicit representation-based method designed for dynamic environments. It eliminates the dynamic objects using estimated masks.
    \item \textsf{PG-SLAM}~\cite{PG-SLAM}: 
    A Gaussian splatting-based method suitable for dynamic environments. It reconstructs dynamic objects online, 
    based on a priori motion constraints.
    \item \textsf{MonST3R}~\cite{zhang2024monst3r}: 
    A geometry-based method that relies on the alignment between point clouds. It can directly output dynamic point clouds. 
    \item \textsf{WildGS-SLAM}~\cite{WildGS}: 
    A Gaussian splatting-based method designed for dynamic environments. It eliminates dynamic objects by learning dynamic regions online.
\end{itemize}
Among them, \textsf{ORB-SLAM3}, \textsf{Rodyn-SLAM}, and \textsf{PG-SLAM} take RGB-D images as input, while \textsf{MonST3R}, \textsf{WildGS-SLAM}, and our \textsf{Dream-SLAM} only use RGB images.

\subsubsection{Localization Results}
Table~\ref{table:com_traj} shows comparisons on both
\textsf{Bonn} and \textsf{TUM} datasets. 
On the \textsf{Bonn} dataset, \textsf{ORB-SLAM3} shows the highest errors.
The reason is that it
cannot tolerate large fractions of matches in dynamic objects. \textsf{Rodyn-SLAM} and \textsf{WildGS-SLAM} achieve higher accuracy by exploiting the constraints of background, but disregard foreground information. \textsf{MonST3R} considers the alignment constraints of both foreground and background, but heavily depends on the pretrained model’s outputs, which may not be reliable in practice. \textsf{PG-SLAM} incorporates both foreground and background for camera localization. However, we empirically observed that it cannot effectively leverage foreground objects with large motions. Our \textsf{Dream-SLAM} achieves the highest accuracy by providing virtual observation constraints, enabling more effective use of foreground information. 

On the \textsf{TUM} dataset, there are several sequences with relatively small camera and human motions. \textsf{ORB-SLAM3} achieves satisfactory performance on these sequences. However, on other highly dynamic sequences, it leads to significant errors. Similarly,
\textsf{MonST3R} becomes unstable on highly dynamic sequences. By contrast, \textsf{RoDyn-SLAM} maintains good stability by filtering out the foreground. \textsf{PG-SLAM}, which leverages both foreground and background information, achieves better performance. 
\textsf{WildGS-SLAM} performs relatively well thanks to its uncertainty map-based optimization.
Despite this, our \textsf{Dream-SLAM}
surpasses it by
incorporating cross-spatio-temporal observations and exploiting both foreground and background information for localization.
\begin{table*}[t]
	\renewcommand{\tabcolsep}{7.15pt}
	\renewcommand\arraystretch{1.3}
	\footnotesize
	\centering
    	\caption{Rendering quality comparison 
        on the \textsf{Bonn}~\cite{ReFusion} and \textsf{TUM}~\cite{tumbenchmark} datasets.
    ``w/o'' and ``w'' represent rendering evaluations without and with considering the dynamic foregrounds, respectively.} 
	\begin{tabular}{cc|cc|cc|cc|cc|cc|cc}
		\toprule
		 & & \multicolumn{6}{c|}{\textsf{Bonn} dataset~\cite{ReFusion}} & \multicolumn{6}{c}{\textsf{TUM} dataset~\cite{tumbenchmark}} \\
		\cmidrule(lr){3-8} \cmidrule(lr){9-14}
		
		  &   & \multicolumn{2}{c}{Seq.~\texttt{ps\_track2}} & \multicolumn{2}{c}{Seq.~\texttt{ps\_track}} & \multicolumn{2}{c|}{Seq.~\texttt{mv\_box2}} & \multicolumn{2}{c}{Seq.~\texttt{wk\_st}} & \multicolumn{2}{c}{Seq.~\texttt{st\_hf}} & \multicolumn{2}{c}{Seq.~\texttt{wk\_xyz}}   \\
		
		\cmidrule(lr){3-8} \cmidrule(lr){9-14}
		  &  & w/o & w  & w/o & w  & w/o & w  &  w/o & w &  w/o & w   & w/o & w    \\
		
		\multirow{3}{*}{\makecell[c]{\textsf{RoDyn-SLAM} \\ \cite{jiang2024rodyn}}} & PSNR$\uparrow$
		& 18.46 & 16.12 & 18.53 & 16.13  & 18.30 & 17.48  & 11.48 & 11.40  & 11.47 & 11.11 & 11.93 & 11.91  \\ 
		& SSIM$\uparrow$  & 0.745 & 0.656 & 0.742 & 0.659  & 0.731 & 0.695  & 0.684 & 0.522  & 0.514 & 0.674 & 0.567 & 0.384   \\
		& LPIPS$\downarrow$ & 0.545 & 0.649 & 0.506 & 0.597  & 0.571 & 0.612  & 0.502 & 0.663  & 0.472 & 0.511 & 0.570 & 0.745  \\
		\midrule
		
		\multirow{3}{*}{\makecell[c]{\textsf{WildGS-SLAM}\\~\cite{WildGS} \\ }} & 
        PSNR$\uparrow$
		& 21.87  & 18.12  & 22,13 & 18.55 & 21.40 & 20.13 & 21.60 & 16.61 & 19.49 & 17.94 & 17.11 &  14.44   \\
		& SSIM$\uparrow$  & 0.827  & 0.791 & 0.817 & 0.787  & 0.783 & 0.770  &  0.844 &  0.7473  & 0.745 & 0.691 & 0.684 & 0.617    \\
		& LPIPS$\downarrow$ & 0.263 & 0.317 & 0.296 &  0.341  & 0.401 &  0.421 &  0.135 & 0.261 & 0.255 &  0.331 &  0.254 & 0.370 \\
        \midrule

        \multirow{3}{*}{\makecell[c]{\textsf{PG-SLAM}\\~\cite{PG-SLAM}}} & PSNR$\uparrow$
		& 26.79 &  26.82  &  26.68 &  27.39 &  \textbf{27.25} &  \textbf{27.38} &  23.55 &  25.65  &  24.96 &  25.46 &  21.56 &  22.99     \\
		& SSIM$\uparrow$  &  0.955  &  0.956 &  \textbf{0.923} &  \textbf{0.958}  &  0.947 &  0.948  &  0.939  &  0.969   &  0.933 &  0.941  &  \textbf{0.937} &  \textbf{0.95}6    \\
		& LPIPS$\downarrow$ &  0.166 &  0.167 &  0.179 &  0.157   &  0.195 &  0.191  &  0.155  &  0.073 &  0.170 &  0.162  &  0.125 &  0.096 \\
        \midrule
        
        \multirow{3}{*}{\makecell[c]{\textsf{Dream-SLAM} \\ (ours)}} & PSNR$\uparrow$
		&  \textbf{29.50} & \textbf{29.69}  & \textbf{27.18} & \textbf{27.45} & 26.60 & 26.71 &  \textbf{25.92} & \textbf{26.26} & \textbf{25.06} & \textbf{25.73} & \textbf{22.42} & \textbf{23.31}    \\
		& SSIM$\uparrow$  &  \textbf{0.957} & \textbf{0.960} & 0.911 & 0.917  & \textbf{0.957} & \textbf{0.958}  &  \textbf{0.964} & \textbf{0.971}   & \textbf{0.942} &  \textbf{0.951} & 0.933 &  0.945   \\
		& LPIPS$\downarrow$ & \textbf{0.077} &  \textbf{0.072}& \textbf{0.158} & \textbf{0.149}   & \textbf{0.097} & \textbf{0.095}  & \textbf{0.091}  & \textbf{0.067} & \textbf{0.106} & \textbf{0.090}  & \textbf{0.095} & \textbf{0.079} \\
		\bottomrule 
		\label{tab:rendering}
	\end{tabular}
	
\end{table*}

\subsubsection{Mapping Results}

\setlength{\tabcolsep}{3.5pt}
\begin{table}[!t]
\renewcommand\arraystretch{1.3}
\renewcommand{\tabcolsep}{27pt} 
\centering
\caption{Efficiency comparison 
in terms of localization and mapping on Sequence \texttt{mv\_box} in the \textsf{Bonn} dataset~\cite{ReFusion}.}
\begin{tabular}{c|c}
\toprule
    & Time Cost Per Frame  \\
    \midrule
    \textsf{WildGS-SLAM}~\cite{WildGS} & 2.79 s\\
    \textsf{PG-SLAM}~\cite{PG-SLAM} & 1.93 s \\
    \textsf{Rodyn-SLAM}~\cite{jiang2024rodyn} & 1.42 s \\
    \textsf{Dream-SLAM} (ours) & \textbf{0.65} s \\
    \bottomrule
\end{tabular}
\label{tab:time_cost}
\end{table}
We quantitatively evaluate the rendering performance, as shown in Table~\ref{tab:rendering}. 
Since some methods cannot reconstruct the foreground, we compute the evaluation metrics based on two settings: over the entire image and with the foreground regions masked out. We also provide qualitative comparison in novel-view renderings, as illustrated in Fig.~\ref{Tum_mapping}. To ensure a fair comparison, the reconstructed scenes of different methods are rendered from the same viewpoint.

\textsf{Rodyn-SLAM} merely focuses on background mapping. As a NeRF-based method, it shows a noticeable gap compared to Gaussian splatting-based approaches. \textsf{WildGS-SLAM} cannot reconstruct the foreground, resulting in lower performance when evaluating the full images. Moreover, its learned uncertainty mask is sensitive to noise in practice, resulting in floating artifacts in background rendering. \textsf{MonST3R} adopts 3D points as reconstruction primitives, thus the discrete projections exhibit unsatisfactory quality. \textsf{PG-SLAM} achieves relatively accurate foreground reconstruction. However, for background regions, depth information may be unstable due to the limited measurement range of the depth camera. As a result, \textsf{PG-SLAM} frequently
produces distorted reconstructions in far fields or boundary areas. Our \textsf{Dream-SLAM} achieves better performance in both foreground and  background reconstructions. The key reason is that it maps the scene by our deep geometry model rather than depth measurements, which avoids the influence of outliers and better preserves the spatial relationships between objects. In addition, our tracking strategy improves reconstruction quality by providing more accurate camera poses.  

\subsubsection{Efficiency Evaluation}
We compare the per-frame time costs of different methods on a representative sequence \texttt{mv\_box} of the \textsf{Bonn} dataset~\cite{ReFusion} that simultaneously involves non-rigid human, rigid box, and static background. 
As shown in Table~\ref{tab:time_cost}, our \textsf{Dream-SLAM} achieves the highest efficiency. 
Among its total runtime of 0.65 s,
dreaming cross-spatio-temporal images accounts for about 0.3 s. \textsf{WildGS-SLAM} is the most time-consuming due to the relatively complex uncertainty map prediction. Both \textsf{Rodyn-SLAM} and \textsf{PG-SLAM} need to train the mapping networks in an online manner, which are slower than our feedforward 3D mapping  network.
In addition, for localization, our \textsf{Dream-SLAM} combines both geometric and photometric constraints, leading to a smaller number of optimization iterations than the other methods.

\subsection{Exploration Planning}

\subsubsection{Experimental Setup}
We introduce datasets, evaluation metrics, implementation details, and methods for comparison.

\textbf{Datasets.} We conduct experiments on the widely used \textsf{Gibson}~\cite{gibsonenv} and \textsf{HM3D}~\cite{ramakrishnan2021hm3d} datasets:
\begin{itemize}
\item  \textsf{Gibson} dataset was established based on the real-world indoor data. These data are processed by the Habitat simulator~\cite{savva2019habitat} to generate observations from arbitrary positions within the scene. 

\item 
\textsf{HM3D} dataset has 
a similar establishment pipeline 
to that of the \textsf{Gibson} dataset. The main difference is that the \textsf{HM3D} dataset features larger and more complicated scenes than the \textsf{Gibson} dataset.
\end{itemize}
The quality of images provided by the above datasets
is relatively low, which makes them unsuitable for localization and mapping evaluation. Please note that these datasets originally do not contain dynamic humans. To evaluate the performance of the algorithms in dynamic environments, we follow~\cite{puig2024habitat} to add dynamic humans into rooms on purpose. We conduct experiments in both dynamic and static environments.

\begin{table*}[!t]
\renewcommand\arraystretch{1.3} 
\centering
\caption{Exploration planning comparison 
on \textsf{Gibson}~\cite{gibsonenv} and \textsf{HM3D}~\cite{ramakrishnan2021hm3d} datasets.
``-'' represents the failure of the full exploration.}
\begin{threeparttable}
\begin{tabular}{c}
\renewcommand{\tabcolsep}{2.25pt}
\renewcommand\arraystretch{1.6}
\begin{tabular}{c|ccc|ccc|ccc|ccc|ccc|ccc||ccc}
            \toprule
            \multicolumn{22}{c}{\textsf{Gibson} dataset~\cite{gibsonenv}}\\
            \midrule
            \multicolumn{1}{c}{Sequences}  & \multicolumn{3}{c}{\texttt{CanWell}} & \multicolumn{3}{c}{\texttt{Eastville}} & \multicolumn{3}{c}{\texttt{Swormville}} & \multicolumn{3}{c}{\texttt{Aloha}} & \multicolumn{3}{c}{\texttt{Nicut}} & \multicolumn{3}{c}{\texttt{Quantico}}  & \multicolumn{3}{c}{Average} \\
				\midrule
				
				\textit{} & CR$^*$↑ & CR↑ & PL↓  & CR$^*$↑  & CR↑ & PL↓ 
                & CR$^*$↑ & CR↑ & PL↓ & CR$^*$↑  &  CR↑ & PL↓ & CR$^*$↑  &  CR↑ & PL↓  & CR$^*$↑ & CR↑ & PL↓  & CR$^*$↑  & CR↑ & PL↓\\

                \textsf{ANM}~\cite{yan2023active} & - &  -  & - 
                & 47.7 & 85.0  & 102.8  
                & 44.5 & 60.5 & 70.9  
                & 32.8 & 82.5 & 90.8  
                & 46.7 &  77.4 & 85.7 
                & 53.0 & 82.1  &  90.6 
                & 44.9 & 77.5 & 88.2\\
                
                \textsf{ANM-S}~\cite{kuang2024active}  
                & 70.3 & 97.0   &  91.8 
                & 64.4 & 91.5  &  70.6 
                & 74.8 & 93.6 & 95.2  
                & 63.6 & 95.9   & 75.8  
                & 72.4 &  97.3 & 83.0 
                & 63.5 & 95.4 & 79.2 
                & 68.1 & 95.1  & 82.6\\
                \textsf{ActiveSplat}~\cite{ActiveSplat}  
                & 64.8 &   97.1  &  116.2 
                & 71.9 &  94.7 & 84.8  
                & 74.7 &  95.3 &  91.9 
                & 71.3 &  91.8  &  82.5 
                & 88.9 &  97.8 &  94.1 
                & 43.6 &  95.1 &  93.7 
                & 69.2 & 95.3 & 93.9\\
                
				\multirow{1}{*}{\makecell[c]{\textsf{Dream-SLAM} \\ (ours)}}   
                & \textbf{98.0} &  \textbf{98.1} & \textbf{87.6} 
                & \textbf{95.4} &  \textbf{95.3} &  \textbf{57.1}  
                & \textbf{95.9}  &  \textbf{95.9}  &  \textbf{51.6} 
                & \textbf{98.2} &  \textbf{98.1} &  \textbf{63.1}  
                & \textbf{98.0} &  \textbf{98.1} &  \textbf{64.4} 
                & \textbf{95.6} &  \textbf{95.6} &  \textbf{62.0} 
                & \textbf{96.8} &  \textbf{96.9} & \textbf{64.3}\\
				\midrule
                \midrule

\end{tabular}
\\
\renewcommand{\tabcolsep}{1.655pt}
\renewcommand\arraystretch{1.6}
    \begin{tabular}{c|ccc|ccc|ccc|ccc|ccc|ccc||ccc}
        
        \multicolumn{22}{c}{\textsf{HM3D } dataset~\cite{ramakrishnan2021hm3d}}\\
            \midrule
            \multicolumn{1}{c}{Sequences}  & \multicolumn{3}{c}{\texttt{CETmJJqkhcK}} & \multicolumn{3}{c}{\texttt{7dmR22gwQpH}} & \multicolumn{3}{c}{\texttt{7UdY7HiDnUi}} & \multicolumn{3}{c}{\texttt{4h4JxvG3cip}} & \multicolumn{3}{c}{\texttt{6HMiy15cxis}} & \multicolumn{3}{c}{\texttt{T7nCRmufFNR}}  & \multicolumn{3}{c}{Average} \\
				\midrule
				
				\textit{} & CR$^*$↑ & CR↑ & PL↓  & CR$^*$↑  & CR↑ & PL↓ 
                & CR$^*$↑ & CR↑ & PL↓ & CR$^*$↑  &  CR↑ & PL↓ & CR$^*$↑  &  CR↑ & PL↓  & CR$^*$↑ & CR↑ & PL↓  & CR$^*$↑  & CR↑ & PL↓\\

				\textsf{ANM}~\cite{yan2023active} 
                & 28.3 & 64.9    & 112.7  
                & - & -  & - 
                & 40.3 &  42.4 & 92.6  
                & 36.8 &  56.0   & 142.3   
                & 39.5 &  61.6 & 144.6 
                & 49.5 & 68.9 & 107.9  
                & 38.9 & 58.7 & 120.0 \\
                
                \textsf{ANM-S}~\cite{kuang2024active}  
                & 33.0 &  92.1  &   97.4 
                & 36.7 & 90.2  & 159.2  
                & 24.1 & 91.9 & 132.1  
                & 49.3 & 93.1 & 99.5 
                & 58.2 & 95.1 & 108.6 
                & 86.9 & 95.4 & 132.2 
                & 48.0 & 92.9 & 121.5\\
                
                \textsf{ActiveSplat}~\cite{ActiveSplat} 
                & 30.6 & 92.9 & 102.9 
                & 81.3 & 94.1 & 180.0 
                & 63.7 & 95.2 & 99.8 
                & 54.3 & 92.2 & 118.8 
                & 76.5 & 95.4 & 114.6 
                & 82.9 & 94.8 & 105.37 
                & 64.9 & 94.2 & 125.8\\
                
				\multirow{1}{*}{\makecell[c]{\textsf{Dream-SLAM} \\ (ours)}}   
                & \textbf{93.2} & \textbf{93.2 } & \textbf{84.7} 
                & \textbf{95.0} & \textbf{95.2} & \textbf{154.7} 
                & \textbf{95.9} & \textbf{96.0} & \textbf{75.8} 
                & \textbf{95.3} & \textbf{95.1} & \textbf{89.6}  
                & \textbf{95.7} & \textbf{95.7} & \textbf{85.9} 
                & \textbf{96.9} & \textbf{96.8} & \textbf{75.8} 
                & \textbf{95.3} & \textbf{95.3} & \textbf{99.4}\\
				\bottomrule 
\end{tabular}
\end{tabular}

\end{threeparttable}
\label{tab:path_plan}
\end{table*}

\textbf{Evaluation Metrics.} For quantitative evaluation, we follow~\cite{ActiveSplat} to compute the total path length (PL) in units of meters, and reconstruction completeness ratio (\%)
defined by the coverage of the reconstructed area against the complete area. We denote this completeness ratio obtained in the dynamic and static environments by CR* and CR, respectively. Please note that we empirically observed that state-of-the-art active SLAM methods fail to explore the entire dynamic environments (reasons are introduced in the following text and supplementary material). Accordingly, PL in dynamic environments is not applicable to these methods, and thus we only report PL in static scenes.
For qualitative analysis, we provide top-view rendering results of the maps, illustrating the explored areas and traversed paths at different exploration progress. 
\begin{figure*}[!t] 
    \centering
    \includegraphics[width=0.92\textwidth]{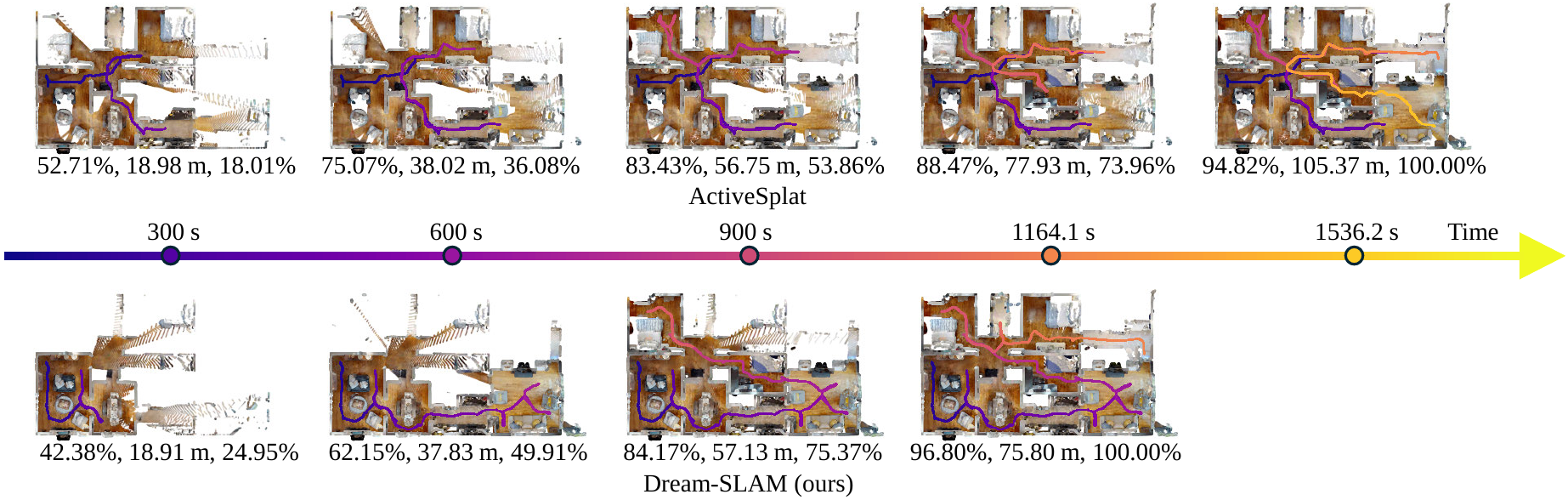} 
        \caption{\textbf{Comparison between \textsf{ActiveSplat}~\cite{ActiveSplat} and our \textsf{Dream-SLAM} in terms of exploration progress on Sequence \texttt{T7nCRmufFNR} of the \textsf{HM3D} dataset~\cite{ramakrishnan2021hm3d}.} We provide a top-view visualization of the mapped scenes at some representative timestamps.
    A triplet of numbers below each image indicates CR, the length of the traversed path, and the proportion of the traversed path to the total path. 
    The trajectory color reflects the time cost of exploration.}
    \label{fig:T7}
\end{figure*}
\begin{figure*}[!t] 
    \centering
    \includegraphics[width=0.92\textwidth]{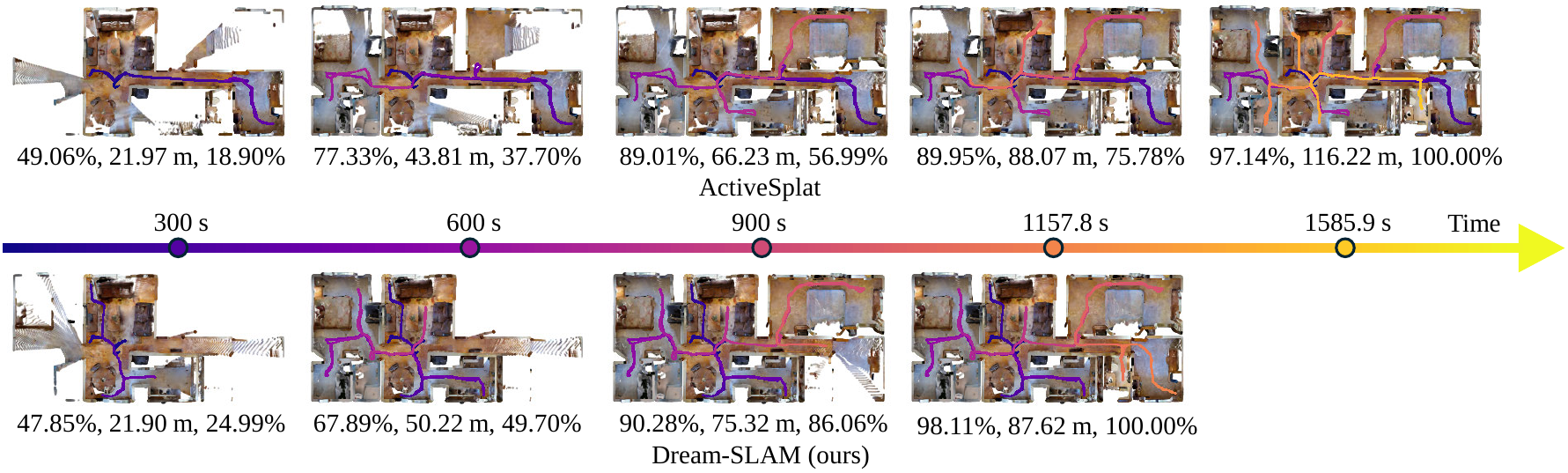} 
    \caption{\textbf{Comparison between \textsf{ActiveSplat}~\cite{ActiveSplat} and our \textsf{Dream-SLAM} in terms of exploration progress on Sequence \texttt{Cantwell} of the \textsf{Gibson} dataset~\cite{gibsonenv}.} We provide a top-view visualization of the mapped scenes at some representative timestamps.
    A triplet of numbers below each image indicates CR, the length of the traversed path, and the proportion of the traversed path to the total path. 
    The trajectory color reflects the time cost of exploration.}
    \label{fig:Cantwell}
\end{figure*}

\textbf{Implementation Details.}  
For the diffusion model to dream structures of unexplored areas (introduced in Section~\ref{sec:path_dreanm}), we initialize it with  
the pretrained Stable Diffusion v2~\cite{rombach2022high} and fine-tune it using LoRA with a rank of 8. To establish the training set for the above fine-tuning, we use the scenes from the \textsf{Matterport3D} dataset~\cite{mp3d}, which do not overlap with the scenes of the \textsf{HM3D} dataset. This practice is helpful to evaluate the generalization when testing on the \textsf{HM3D} and \textsf{Gibson} datasets. We set the pixel opacity threshold $\tau_\textnormal{o}$ to 0.5, and set the occupancy score thresholds $\underline{\tau}$ and $\overline{\tau}$ to 0.2 and 0.5, respectively  (see Section~\ref{sec:plan_gobal_Predict}). We empirically observed that our method is robust to these hyperparameters.

\textbf{Methods for Comparison.} 
We compare the planning module of our \textsf{Dream-SLAM} against the following state-of-the-art active SLAM methods introduced in Section~\ref{sec:rel_work}:
\begin{itemize}
   
    \item \textsf{ANM}~\cite{yan2023active}: 
    A neural representation-based method that determines the target goal by minimizing map uncertainty. It is a classic greedy strategy.

     \item \textsf{ANM-S}~\cite{kuang2024active}: 
    An implicit representation-based method that is an improvement of the above \textsf{ANM}. It performs planning based on a generalized Voronoi graph.
    
    \item \textsf{ActiveSplat}~\cite{ActiveSplat}: 
    A Gaussian splatting-based method that is an improvement of the above \textsf{ANM-S}. It prioritizes the exploration of local regions.
    
\end{itemize}

\setlength{\tabcolsep}{3.5pt}
\begin{table}[!t]
\renewcommand\arraystretch{1.3}
\renewcommand{\tabcolsep}{8.0pt} 
\centering
\caption{Ablation study of localization using the dreamed cross-spatio-temporal images.}
\begin{tabular}{c|cc|cc}
			\toprule
			\multicolumn{1}{c}{Sequence}  &  \multicolumn{2}{c}{Without  Dreaming}& \multicolumn{2}{c}{Dreaming (ours)}  \\
   
		\midrule
                \textit{}  & RMSE$\downarrow$ & SD$\downarrow$  & RMSE$\downarrow$ & SD$\downarrow$ \\
			 \texttt{wk\_hf} of \textsf{TUM}~\cite{tumbenchmark}  & 1.8 & 0.7 & \textbf{1.6}   & \textbf{0.7}   \\
                \texttt{ps\_track} of \textsf{Bonn}~\cite{ReFusion}  &  1.7& 0.7  & \textbf{1.5}   & \textbf{0.5}  \\
			\bottomrule 
			
		\end{tabular}
		\label{tab:ab_localize_dream}
		
\end{table}
\begin{figure}[!t]
\footnotesize
	\renewcommand\arraystretch{1.2} 
	\renewcommand{\tabcolsep}{1pt} 
	\centering
\begin{tabular}{cc}
Without Dreaming & Dreaming (ours) \\
 \includegraphics[width=0.45\linewidth]{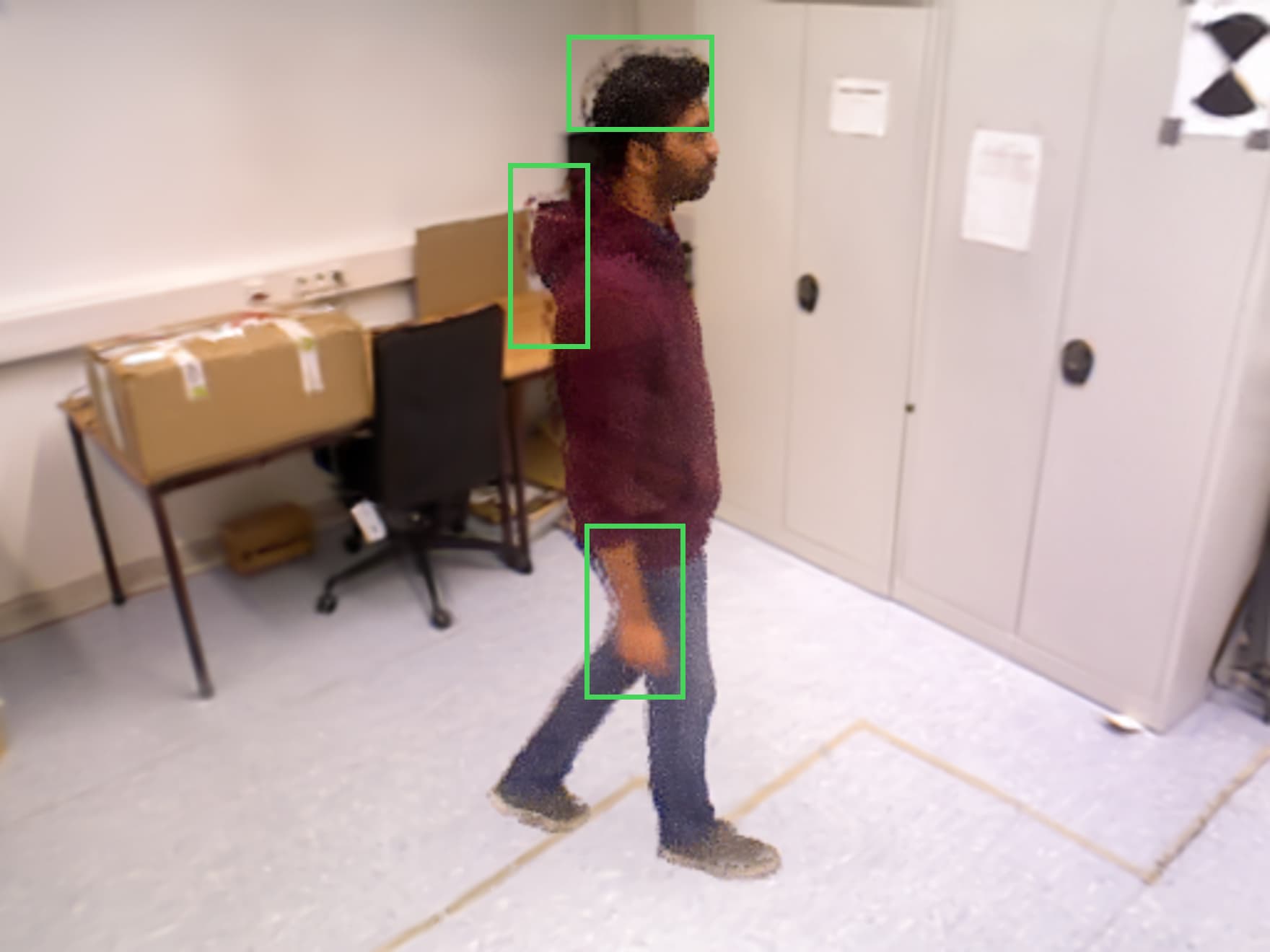}  & \includegraphics[width=0.45\linewidth]{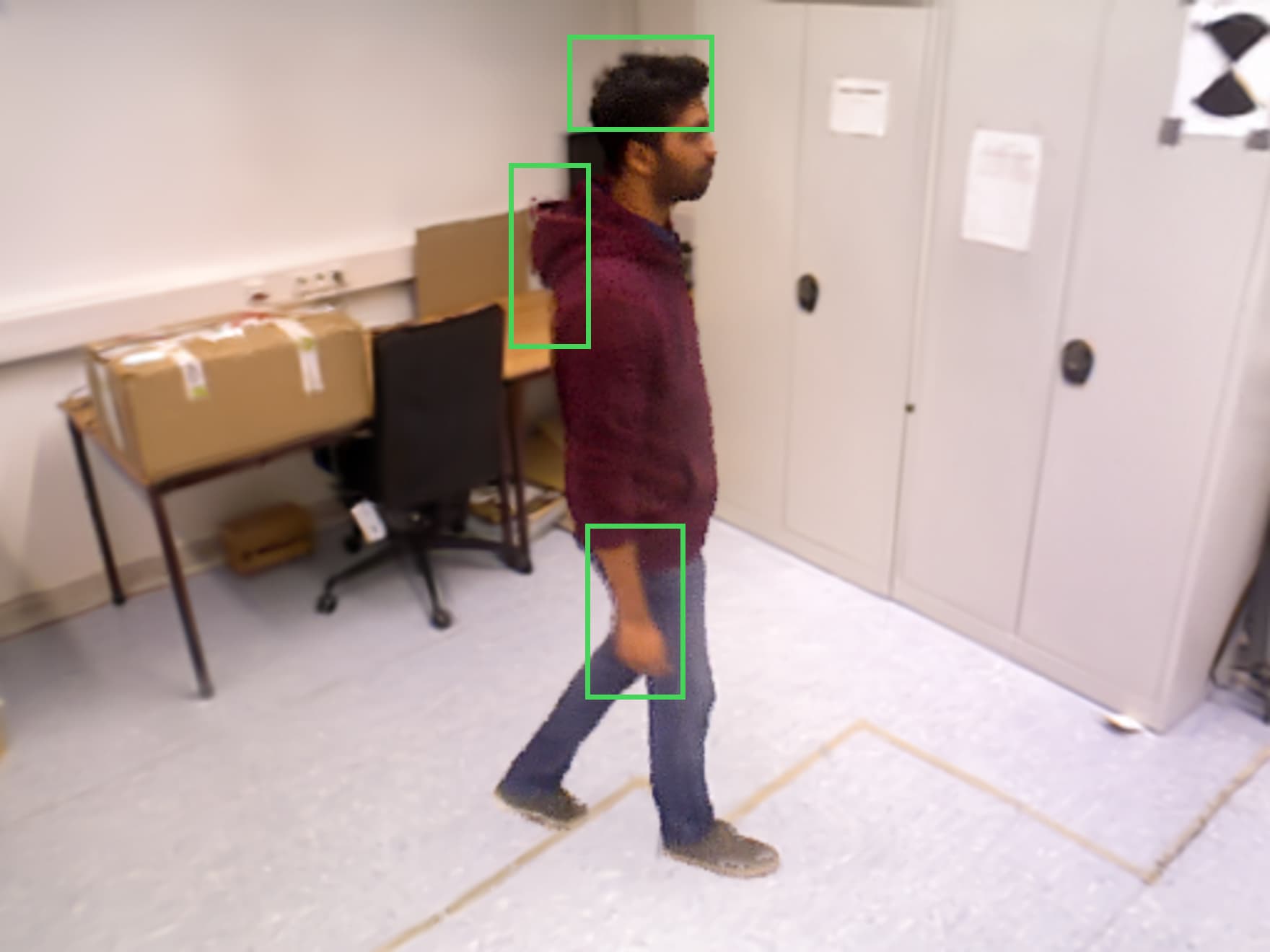}  \\
\multicolumn{2}{c}{(a)} \\[0.5em]
 \includegraphics[width=0.45\linewidth]{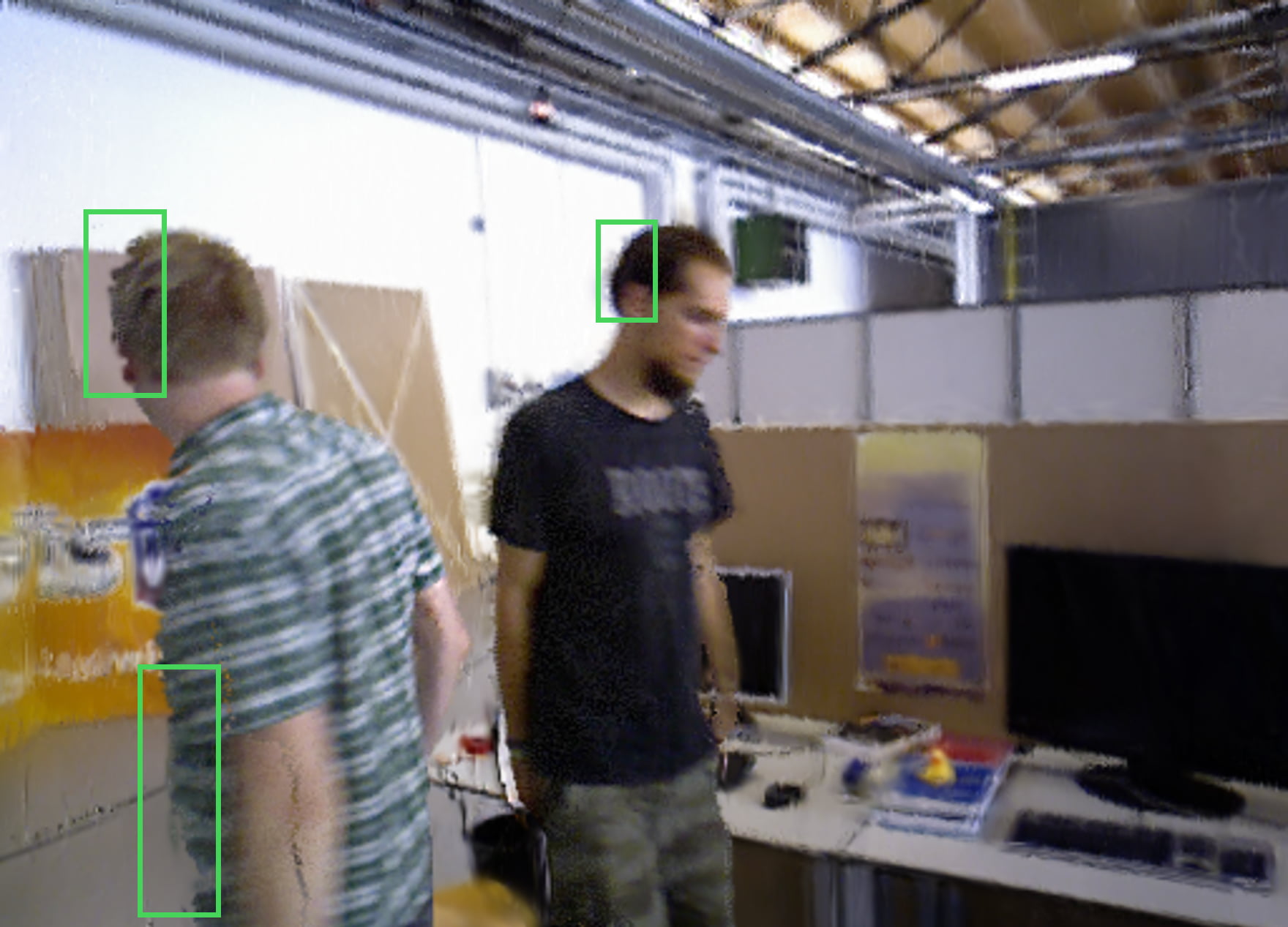}  & \includegraphics[width=0.45\linewidth]{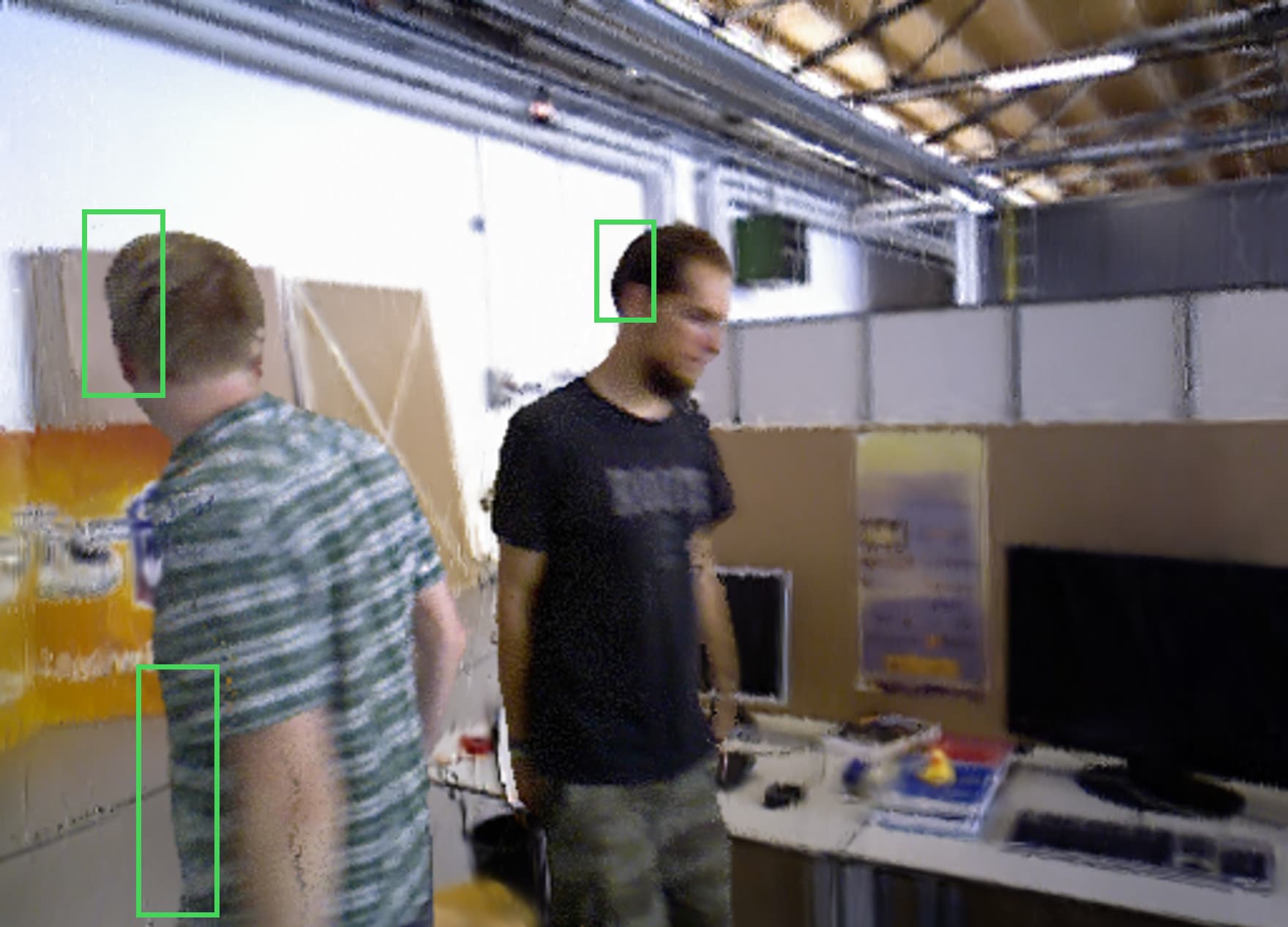}  \\
\multicolumn{2}{c}{(b)}
\end{tabular}
\caption{\textbf{Ablation study of mapping using the dreamed cross-spatio-temporal images.}  We compare the rendering results of Gaussian maps obtained without and with the dreamed images on (a) Sequence \texttt{ps\_track} of the \textsf{Bonn} dataset~\cite{ReFusion}, and (b) Sequence \texttt{wk\_hf} of the \textsf{TUM} dataset~\cite{tumbenchmark}.}
       	\label{ablation_mapping}
\end{figure}

\subsubsection{Evaluation Results}
The quantitative evaluations in both dynamic and static environments are presented in Table~\ref{tab:path_plan}. 
We also provide qualitative visualization of the exploration process in static environments,
as shown in Figs.~\ref{fig:T7} and~\ref{fig:Cantwell}. 
Overall, existing methods fail to explore the entire dynamic scenes, leading to CR* significantly smaller than CR. The main reason is that they typically treat dynamic humans as obstacles that occlude the doorways and entrances to other rooms. By contrast, our \textsf{Dream-SLAM} can map dynamic humans and static background separately, and thus is not affected by the dynamic humans during planning.
We provide a more detailed analysis in terms of CR and PL as follows. On both datasets, \textsf{ANM} exhibits unsatisfactory performance since its waypoint selection strategy based on map uncertainty tends to overlook certain areas. \textsf{ANM-S} achieves relatively high exploration completeness but with longer paths. Its Voronoi graph–based planning strategy prioritizes the waypoints with high uncertainty, resulting in a lack of global planning and frequent backtracking. While \textsf{ActiveSplat} achieves higher exploration completeness by visiting all the waypoints within each sub-region, its path cost further increases as a result.
Its decisions tend to be locally optimal when dealing with multi-connected waypoints. By contrast, our \textsf{Dream-SLAM} provides the highest exploration completeness and the lowest path cost across all sequences of both datasets, benefiting from our farsighted planning that reasons over the semantically plausible structures dreamed for unexplored areas. 

\setlength{\tabcolsep}{3.5pt}
\begin{table}[!t]
\renewcommand\arraystretch{1.3}
\renewcommand{\tabcolsep}{6pt} 
\centering
\caption{Ablation study of planning using the dreamed structures of unexplored areas.}
\begin{tabular}{c|cc|cc}
    \toprule
    \multicolumn{1}{c}{Sequence}  &  \multicolumn{2}{c}{Without  Dreaming} & \multicolumn{2}{c}{Dreaming (ours)}  \\
   
	\midrule
    \textit{}   & CR↑ & PL↓  & CR↑ & PL↓ \\
    
    \texttt{Cantwell} of \textsf{Gibson}~\cite{gibsonenv}  &  
    97.5 & 101.3 &   \textbf{98.1} &  \textbf{88.4} \\
    
    \texttt{T7nCRmufFNR} of \textsf{HM3D}~\cite{ramakrishnan2021hm3d}  &  95.5 & 85.1 &  \textbf{96.8}  & \textbf{75.8}  \\
    
    \bottomrule 
    \end{tabular}
\label{tab:ab_plan_dream}
		
\end{table}
\begin{figure}[!t]
\footnotesize
\centering
\begin{tabular}{cc}
    \includegraphics[height=0.68\linewidth, valign=c]{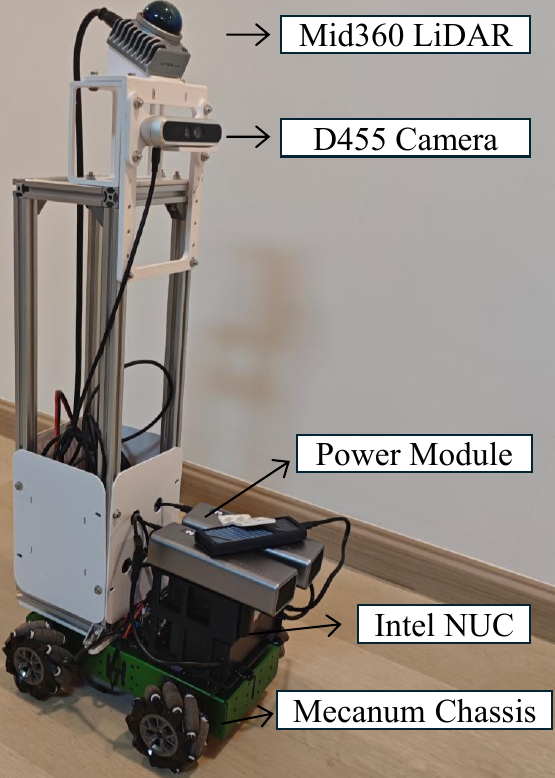} 
    & 
    \begin{tabular}{c}
        \includegraphics[width=0.29\linewidth, valign=c]{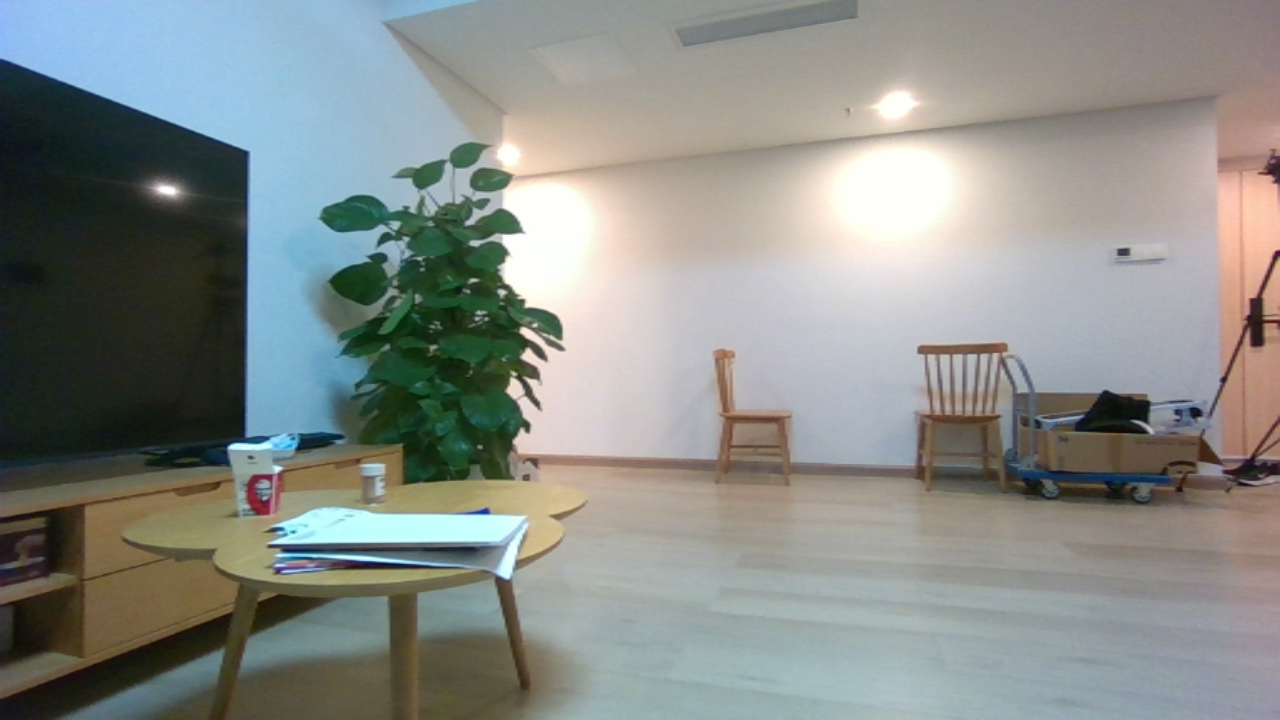} \\
        \noalign{\vskip 2pt} 
        \includegraphics[width=0.29\linewidth, valign=c]{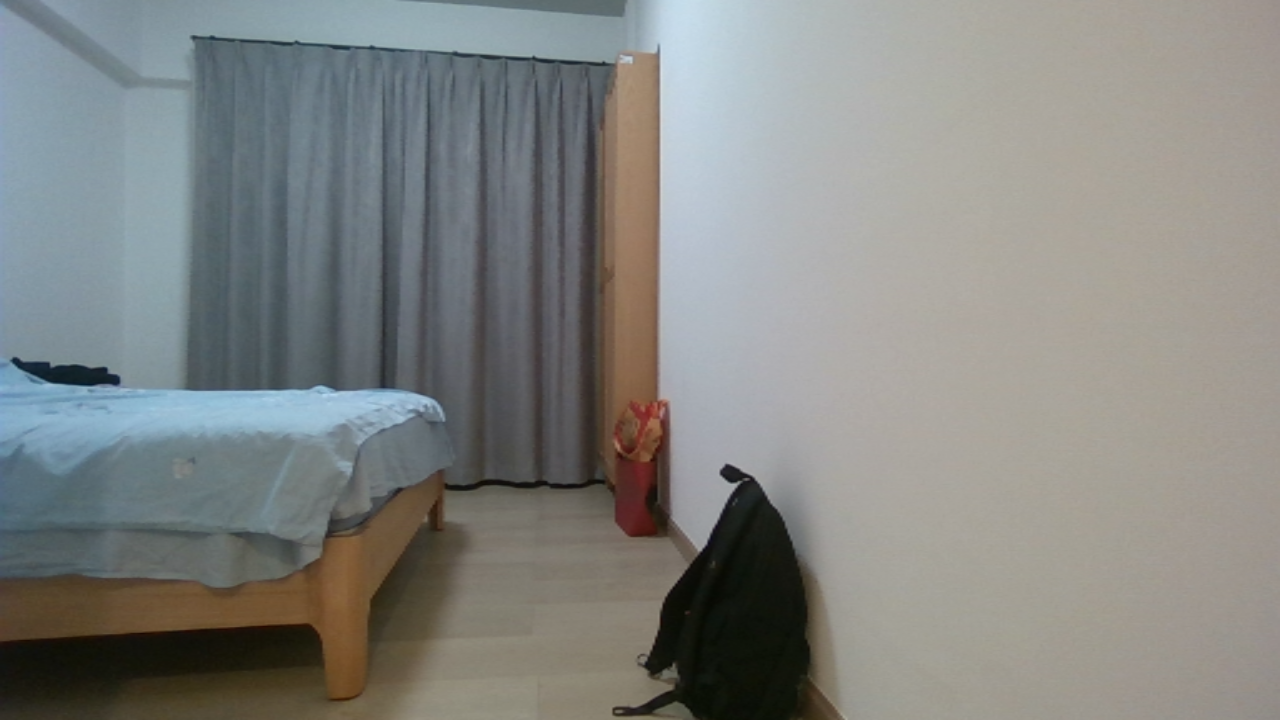} \\
        \noalign{\vskip 2pt}
        \includegraphics[width=0.29\linewidth, valign=c]{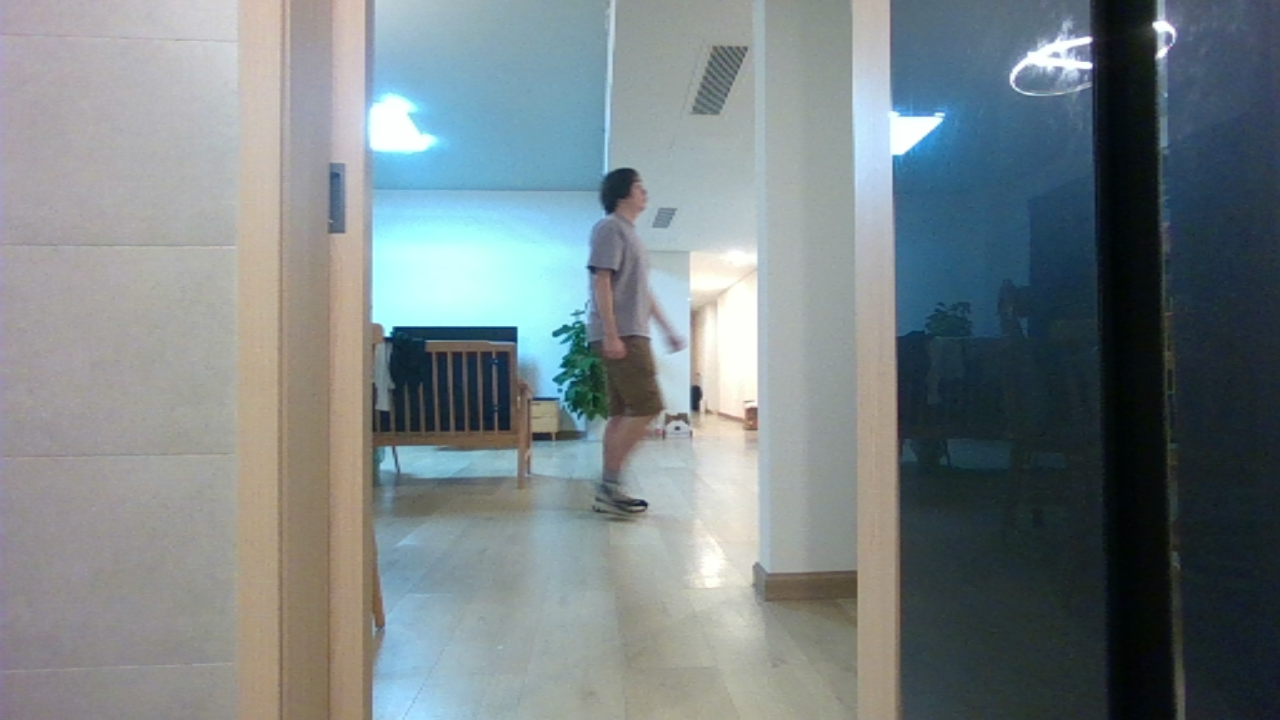} \\
        \noalign{\vskip 2pt}
        \includegraphics[width=0.29\linewidth, valign=c]{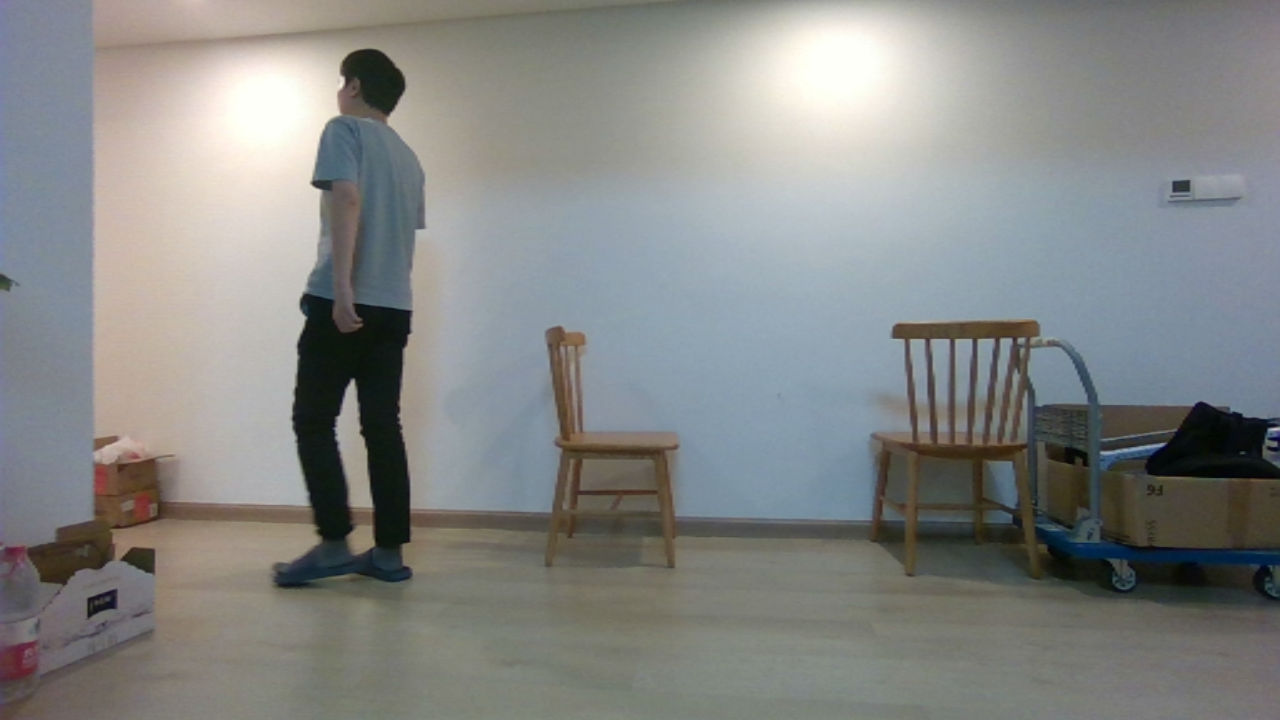}
    \end{tabular}
    \\(a) &  (b)
\end{tabular}
\caption{\textbf{Setup of experiments on our self-collected data.} (a) 
Mobile robot to
collect data. 
It is built on a four-wheeled Mecanum chassis and equipped with a front-facing camera and a LiDAR. (b) Sample images of our self-collected data in a home environment containing dynamic humans.}
\label{fig:robot}
\end{figure}

\subsection{Ablation Study}
We conduct ablation study to validate the effectiveness of the proposed modules and strategies.

\subsubsection{Localization Using Dreamed Images}
\label{subsubsec:Ablation:Foreground}
Recall that for camera pose estimation, we leverage the foreground constraints provided by the dreamed cross-spatio-temporal images. We compare our strategy to the version without using such images.
As shown in Table~\ref{tab:ab_localize_dream}, our strategy achieves higher accuracy. This validates that our image dreaming is reliable, and also the foreground information can effectively enhance the localization accuracy.

\subsubsection{Mapping Using Dreamed 
Images}
Recall that for Gaussian map refinement, we 
incorporate the dreamed cross-spatio-temporal images as a supplement to real observations.
We compare our strategy to the version without using such images.
As shown in Fig.~\ref{ablation_mapping}, we report the qualitative results of dynamic map rendering.\footnote{Due to the lack of ground-truth structures of dynamic environments, a quantitative evaluation is infeasible.}
Results demonstrate that cross-spatio-temporal images can provide additional constraints on foreground reconstruction, and thus reconstruct a more spatially reasonable map.

\subsubsection{Planning Using Dreamed  Structures}
Recall that we plan the path by dreaming 
semantically plausible structures of unexplored areas as a supplement to the observed structures.
For validation, we compare our method to the version without the dreamed structures. 
As shown in Table~\ref{tab:ab_plan_dream},
reasoning over semantically plausible structures enables thorough exploration with reduced path length, thereby yielding superior exploration efficiency.

\section{Experiments on Self-collected Data}

To jointly evaluate localization and mapping, together with exploration planning, we use the self-collected data obtained in the real world.

\begin{figure*}[!t] 
    \centering
    \includegraphics[width=0.95\textwidth]{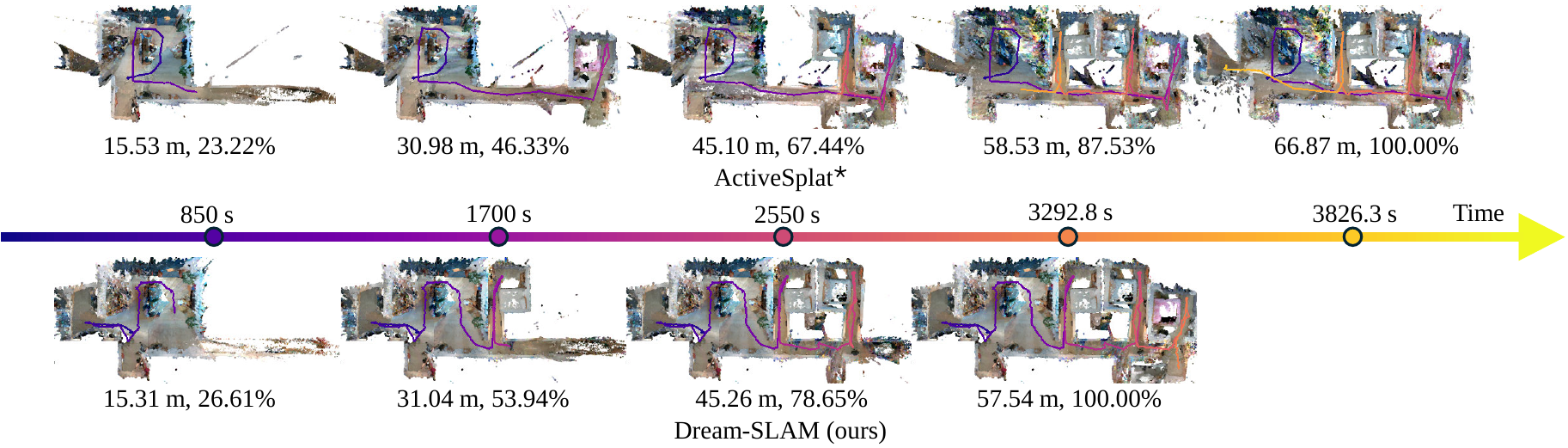} 
    \caption{\textbf{Comparison between \textsf{ActiveSplat*}~\cite{WildGS,ActiveSplat} and our \textsf{Dream-SLAM} in terms of exploration progress on our self-collected data.} We provide a top-view visualization of the mapped scenes at some representative timestamps.
    A pair of numbers below each image indicates
     the length of the traversed path, and the proportion of the traversed path to the total path. 
    The trajectory color reflects the time cost of exploration.}
    \label{fig:real_topview}
\end{figure*}
\subsubsection{Experimental Setup}
As shown in Fig.~\ref{fig:robot}(a), we develop our mobile robot based on a four-wheeled Mecanum chassis. 
The robot integrates 
a front-facing D455 camera to obtain RGB images
and a Livox Mid-360 LiDAR to acquire ground-truth robot trajectories. 
Our robot is not equipped with dedicated graphical computing resources. Instead, computation is performed on a server with an NVIDIA RTX 4090 GPU connected via a local network. For robot movement, we set the maximum linear speed to 0.2 m/s and the maximum angular speed to 0.2 rad/s. As shown in Fig.~\ref{fig:robot}(b), we collect data in a home, which includes a living room, a kitchen, three bedrooms, and two bathrooms. The home contains several humans moving around. 

\begin{figure}[!t]
\footnotesize
\centering
\renewcommand{\tabcolsep}{0.9pt}
\begin{tabular}{ccc} 
    \multicolumn{3}{c}{\includegraphics[width=0.8\linewidth]{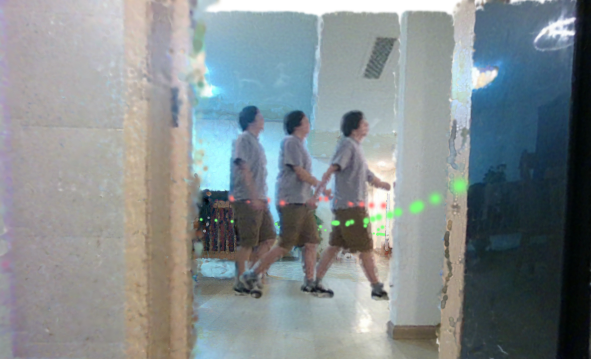}} \\ 
    
\end{tabular}
\caption{\textbf{Dynamic scene mapping of our \textsf{Dream-SLAM} on our self-collected data.}
We render the reconstructed map composed of both static background and dynamic foreground from a novel view. 
The red and green dotted lines denote the trajectories of human and camera, respectively.}
\label{fig:real-world_mapping}
\end{figure}
\setlength{\tabcolsep}{3.5pt}
\begin{table}[!t]
\renewcommand\arraystretch{1.3}
\renewcommand{\tabcolsep}{20.8pt} 
\centering
\caption{Camera localization comparison between \textsf{WildGS-SLAM}~\cite{WildGS} and our~\textsf{Dream-SLAM} on our self-collected data.}
\begin{tabular}{cc|cc}
    \toprule
  \multicolumn{2}{c}{\textsf{WildGS-SLAM}~\cite{WildGS}  }& \multicolumn{2}{c}{\textsf{Dream-SLAM} (ours)}  \\
   
	\midrule
    RMSE$\downarrow$ & SD$\downarrow$  & RMSE$\downarrow$ & SD$\downarrow$ \\
	 16.6 & 7.3 &   \textbf{10.8} & \textbf{5.6}   \\
	\bottomrule 
			
		\end{tabular}
		\label{tab:real_localize}
		
\end{table}

We choose the state-of-the-art active SLAM method \textsf{ActiveSplat} as the comparative approach.
As mentioned above, the original \textsf{ActiveSplat} uses the ground-truth camera poses provided by the simulator, and performs mapping by SplaTAM~\cite{splatam} that is only suitable for static environments. For a fair comparison, we integrate \textsf{ActiveSplat} with the aforementioned state-of-the-art localization method \textsf{WildGS-SLAM} applicable to dynamic scenes. We denote this integration by \textsf{ActiveSplat*}.
Both \textsf{ActiveSplat*} and our \textsf{Dream-SLAM} perform exploration planning from the same starting point. In addition, we independently evaluate the localization by comparing \textsf{WildGS-SLAM} and our \textsf{Dream-SLAM}. For an unbiased comparison, both methods use the same pre-recorded image sequence. This sequence is obtained along the path planned by our \textsf{Dream-SLAM}. We adopt the metrics introduced in Section~\ref{exp_real} for quantitative evaluation.

\subsubsection{Experimental Results}

Fig.~\ref{fig:real_topview} presents the comparison in terms of exploration planning.
To complete the full exploration, our \textsf{Dream-SLAM} travels a total distance of 57.54 m, saving 14\% than \textsf{ActiveSplat*}. Specifically, \textsf{ActiveSplat*} can only exploit the currently observed information and is prone to generating a locally optimal path. For example, it overlooks the kitchen at the beginning and fails to detect several bedrooms later. By contrast, by reasoning over semantically plausible structures imagined for unexplored regions, our \textsf{Dream-SLAM} plans a more farsighted path. This superior performance validates the effectiveness of our strategy and its adaptability to real-world environments. Moreover, Fig.~\ref{fig:real-world_mapping} shows that during planning, our \textsf{Dream-SLAM} can reconstruct the dynamic foreground at different times. This avoids the occlusion of the entrances and doorways to the other rooms. In addition, the independent comparison in terms of camera localization is shown in Table~\ref{tab:real_localize}. Our \textsf{Dream-SLAM} significantly outperforms \textsf{WildGS-SLAM}. With more accurate camera poses, our \textsf{Dream-SLAM} is better equipped to perform downstream path planning.

\section{Conclusions}
In this paper, we propose Dream-SLAM, a monocular active SLAM method via dreaming cross-spatio-temporal images and semantically plausible structures of unexplored areas in dynamic environments. For localization, our method uses the dreamed images to formulate novel constraints, simultaneously exploiting dynamic foreground and static background cues. For mapping, we introduce a feedforward network to predict Gaussians and also use the dreamed images for Gaussian refinement, achieving a photo-realistic and coherent representation of both foreground and background. For exploration planning, we propose a strategy that integrates the dreamed structures with the observed geometry, enabling long-horizon reasoning and producing farsighted trajectories. Extensive experiments on public and self-collected datasets demonstrate that our method outperforms state-of-the-art approaches.

\balance
\bibliographystyle{IEEEtran}
\bibliography{papers}

\end{document}